%% file: main.tex
\documentclass[10pt,twocolumn,letterpaper]{article}

\usepackage{iccv}
\usepackage{times}
\usepackage{epsfig}
\usepackage{graphicx}
\usepackage{amsmath}
\usepackage{amssymb}
\usepackage{epstopdf}
\usepackage{paralist}
\usepackage{bm}
\usepackage{mathtools} 
\usepackage{makecell}
\usepackage{subfigure}
\usepackage[pagebackref=true,breaklinks=true,letterpaper=true,colorlinks,bookmarks=false]{hyperref}

\iccvfinalcopy % *** Uncomment this line for the final submission

\def\iccvPaperID{4041} % *** Enter the ICCV Paper ID here

\ificcvfinal\fi

\begin{document}

%%%%%%%%% TITLE
\title{DiffusionDepth: Diffusion-based Self-Refinement Approach for Monocular Depth Estimation }
\title{DiffusionDepth: Diffusion Denoising Approach for Monocular Depth Estimation }

\author{Yiqun Duan~\textsuperscript{1,$\star$}\\
%{\tt\small yiqun.duan@student.uts.edu.au}
\and
Xianda Guo~\textsuperscript{2,$\ast$}\\
\and
Zheng Zhu~\textsuperscript{2}\\
%\and
%Chin-Teng Lin
}

\maketitle
\ificcvfinal\thispagestyle{empty}\fi

\footnotetext{~\textsuperscript{1} CIBCI Lab, Australia Artificial Intelligence Institute (AAII), University of Technology Sydney, yiqun.duan@student.uts.edu.au }
\footnotetext{~\textsuperscript{2} PhiGent Robotics, \{xianda.guo, zheng.zhu\}@phigent.ai}
\footnotetext{~\textsuperscript{$\star$} First author, ~\textsuperscript{$\ast$} Corresponding Author}
%%%%%%%%% ABSTRACT
\begin{abstract}
Monocular depth estimation is a challenging task that predicts the pixel-wise depth from a single 2D image. Current methods typically model this problem as a regression or classification task. We propose DiffusionDepth, a new approach that reformulates monocular depth estimation as a denoising diffusion process. It learns an iterative denoising process to `denoise' random depth distribution into a depth map with the guidance of monocular visual conditions. The process is performed in the latent space encoded by a dedicated depth encoder and decoder. Instead of diffusing ground truth (GT) depth, the model learns to reverse the process of diffusing the refined depth of itself into random depth distribution. This self-diffusion formulation overcomes the difficulty of applying generative models to sparse GT depth scenarios. The proposed approach benefits this task by refining depth estimation step by step, which is superior for generating accurate and highly detailed depth maps. Experimental results on KITTI and NYU-Depth-V2 datasets suggest that a simple yet efficient diffusion approach could reach state-of-the-art performance in both indoor and outdoor scenarios with acceptable inference time. Codes are available through link~\textsuperscript{3}.
% Detailed ablation results give an intuitive reference to extending diffusion formation for more 3D vision tasks. 
% Codes are available through an anonymous link\footnote{\href{github.diffusiondepth.com}{github.diffusiondepth.com}} 
%Recent approaches improve depth estimation performance by introducing auxiliary branches such as classification-regression branches or uncertainty branches. 
% However, current methods 这里想一句
%However, the clarity of the depth map produced by current methods is not satisfactory as these methods mostly perform merging on low-resolution feature maps and interpolate the feature map to desired inputs. 
%This limited the performance of scenery details. 
%上面也可以没有
%We propose DiffusionDepth, a new framework that formulates object detection as a denoising diffusion process
%from noisy boxes to object boxes. During training stage,object boxes diffuse from ground-truth boxes to random distribution, and the model learns to reverse this noising process.
% URCDC

% reformulate the task as classification-regression or utilizing 

\end{abstract}

%%%%%%%%% BODY TEXT
\footnotetext{~\textsuperscript{3} \href{https://github.com/duanyiqun/DiffusionDepth}{https://github.com/duanyiqun/DiffusionDepth}}

\section{Introduction}

Monocular depth estimation is a fundamental vision task with numerous applications such as autonomous driving, robotics, and augmented reality. 
% Benefiting from the advances in convolutional neural networks (CNNs)~\cite{He_2016_CVPR, tan2019efficientnet}, recent studies~\cite{lee2019big, bhat2021adabins} achieve promising depth results. %where the encoder-decoder-based architecture is leveraged.
Along with the rise of convolutional neural networks (CNNs)~\cite{He_2016_CVPR,tan2019efficientnet,duan2019learning}, numerous mainstream methods employ it as dense per-pixel regression problems, such as RAP~\cite{zhang2019pattern}, DAV~\cite{huynh2020dav}, and BTS~\cite{lee2019bts}.
Follow-up approaches such as UnetDepth~\cite{guo2018learning}, CANet~\cite{yan2021channel}, and BANet~\cite{aich2021bidirectional}, concentrate on enhancing the visual feature by modifying the backbone structure. 
Transformer structures~\cite{vaswani2017attention,dosovitskiy2020vit,liu2021swin,yuan2021tokens} is introduced by DPT~\cite{ranftl2021dpt}, and PixelFormer~\cite{agarwal2023attention} pursue the performance to a higher level by replacing CNNs for better visual representation.
% 这句可以删掉
However, pure regression methods suffer from severe overfitting and unsatisfactory object details.

%-------------------------------------------------------------------------
\begin{figure}[t]
\begin{center}
% \fbox{\rule{0pt}{1.5in} \rule{0.9\linewidth}{0pt}}
   \includegraphics[width=1\linewidth]{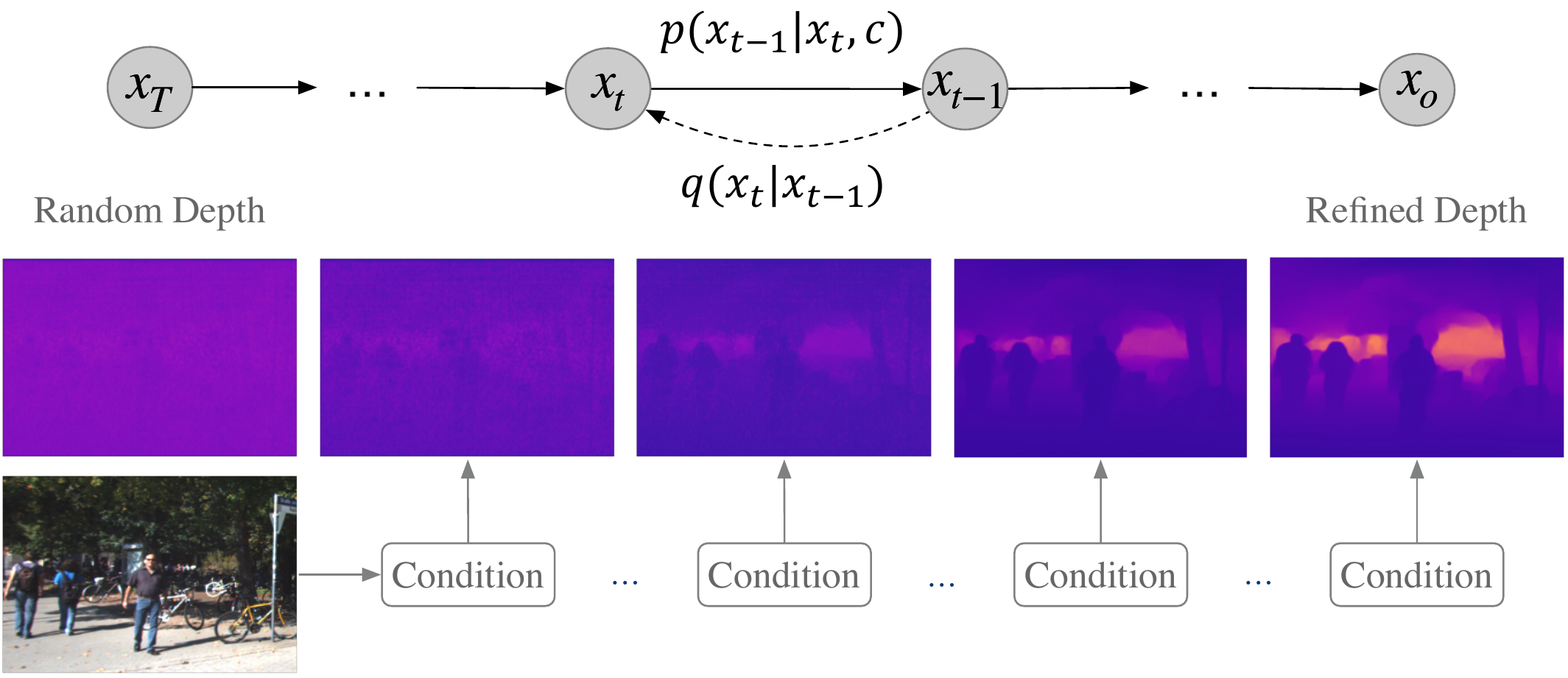}
\end{center}
   %\caption{Illustration of DiffusionDepth, the model refines the depth map $x_t$ with monocular guidance $c$ from random depth initialization $x_T$ to refined estimation result $x_o$. }
   \caption{Illustration of DiffusionDepth, the model refines the depth map $x_t$ with monocular guidance $c$ from random depth initialization $x_T$ to the refined estimation result $x_o$. }
\label{fig:cover}
\end{figure}
%-------------------------------------------------------------------------

Estimating depth from a single image is challenging due to the inherent ambiguity in the mapping between the 2D image and the 3D scene. 
To increase the robustness, the following methods utilizing constructed additional constraints such as uncertainty (UCRDepth~\cite{shao2023urcdc}), and piecewise planarity prior (P3Depth~\cite{patil2022p3depth}). 
The NewCRFs~\cite{DBLP:journals/corr/abs-2203-01502} introduces window-separated Conditional Random Fields (CRF) to enhance local space relation with neighbor pixels. 
DORN~\cite{fu2018deep}, and Soft Ordinary~\cite{diaz2019soft} propose to discretize continuous depth into several intervals and reformulate the task as a classification problem on low-resolution feature maps. 
Follow-up methods (AdaBins~\cite{bhat2021adabins,johnston2020self}, BinsFormer~\cite{li2022binsformer}) merge regression results with classification prediction from bin centers. 
% BinsFormer~\cite{li2022binsformer} improves AdaBins by introducing multi-scale bins supervision and additional chamfer loss~\cite{wu2021density}. 
However, the discretization depth values from bin centers result in lower visual quality with discontinuities and blur.
% However, the discretization classified depth value lead to visual discontinuities. 
% Also, the pixel-to-pixel prediction leads to a limited perception field. 

%as the merging requires bi-linear upsampling values into depth prediction. 

We solve the depth estimation task by reformulating it as an iterative denoising process that generates the depth map from random depth distribution. 
The brief process is described in Fig.~\ref{fig:cover}. 
% Intuitively, the step-by-step refinement could largely improve the visual and prediction quality of small details in the distance by introducing strong generative ability. 
Intuitively, the iterative refinement enables the framework to capture both coarse and fine details in the scene at different steps. 
Meanwhile, by denoising with extracted monocular guidance on large latent space, this framework enables accurate depth prediction in high resolution. 
Diffusion models have shown remarkable success in generation tasks~\cite{hoogeboom2022equivariant, trippe2022diffusion}, or more recently, on detection~\cite{chen2022diffusiondet} and segmentation~\cite{chen2022generalist,chen2022diffusiondet} tasks.
To the best of our knowledge, this is the first work introducing the diffusion model into depth estimation.

\begin{figure*}[hbtp]
\begin{center}
\includegraphics[width=1\linewidth]{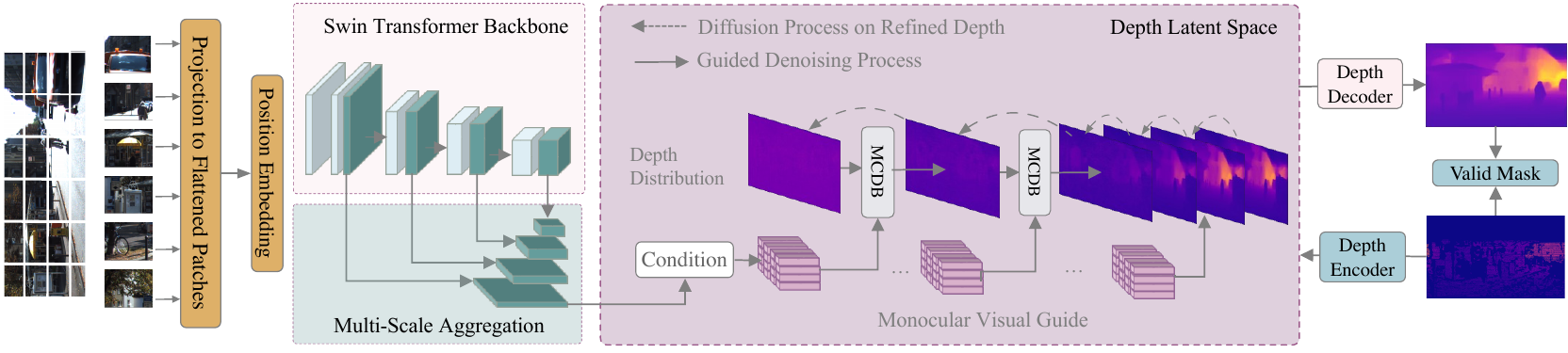}
\end{center}
   %\caption{Overview of DiffusionDepth. Given monocular visual input, the model employs a backbone module to extract multi-scale features. The multi-scale features are aggregated through attention and feature pyramids neck (Multi-Scale FPN) to construct visual guidance conditions. The diffusion head iteratively refines (denoise) the depth distribution from noise initialization to refined depth prediction under the guidance of monocular visual conditions. The guided denoising is realized by the proposed monocular conditioned denoising block (MCDB) as shown above.}
   \caption{\textbf{Overview of DiffusionDepth.} Given monocular visual input, the model employs a feature extractor and multiscale feature aggregation to construct visual guidance conditions. The Monocular Conditioned Denosing Block (MCDB) iteratively refines the depth distribution from noise initialization to refined depth prediction under the guidance of monocular visual conditions. }
\label{fig:mainstructure}
\end{figure*}

% This paper proposes DiffusionDepth, a novel framework tackling monocular estimation tasks by formulating it as an iterative denoising process from random depth distribution given the monocular visual guidance. 
This paper proposes DiffusionDepth, a novel framework for monocular depth estimation as described in Fig.~\ref{fig:mainstructure}. 
%that formulates the task as an iterative denoising process. 
% Specifically, the framework takes a random depth distribution as input and refines it through a series of denoising steps, utilizing monocular visual conditions as guidance. 
The framework takes in a random depth distribution as input and iteratively refines it through denoising steps guided by visual conditions. 
By performing the diffusion-denoising process in latent depth space~\cite{rombach2022high}, DiffusionDepth is able to achieve more accurate depth estimation with higher resolution. 
The depth latent is composed of a subtle encoder and decoder.
%which also enables soft depth value range control for the prediction.
% To perform accurate depth estimation with higher resolution, we perform the diffusion-denoising process in latent depth space. 
% This is achieved by introducing a depth encoder and decoder to encode depth maps into latent space~\cite{rombach2022high}. 
% It also enables soft depth value range control for the prediction.
% The denoise guidance (monocular condition) is constructed with multi-scale visual clues. 
The denoising process is guided by visual conditions by merging it with the denoising block through a hierarchical structure (Fig.~\ref{fig:mcdb}). 
The visual backbone extracts multi-scale features from monocular visual input and aggregated it through a feature pyramid (FPN~\cite{lin2017feature}).
%and hierarchical aggregation and heterogeneous interaction (HAHI~\cite{li2022depthformer}) neck. 
We aggregated both global and local correlations to construct a strong monocular condition. 
% Since the depth latent space is with higher resolution, the merging could avoid direct upsampling from visual features, thus increasing the visual continuity. 

%One severe problem of adopting generative methods into depth prediction is the sparse ground truth (GT) depth value problem, where in most outdoor scenarios, only partial pixels~\footnote{In datasets such as KITTI depth, only a small percentage of pixels ($3.75-5\%$) have GT depth values.} have GT depth value. 
%This sparsity can lead to mode collapse in normal generative training.

One severe problem of adopting generative methods into depth prediction is the sparse ground truth (GT) depth problem~\footnote{In datasets such as KITTI Depth, only a small percentage of pixels ($3.75-5\%$) have GT depth values.}, which can lead to mode collapse in normal generative training.
To address this issue, DiffusionDepth introduces a self-diffusion process. During training, instead of directly diffusing on sparse GT depth values, the model gradually adds noise to refined depth latent from the current denoising output.
%At the training stage, instead of directly diffusing on sparse GT depth, we gradually add noise to refined depth prediction results from the current denoising output. 
% This noise serves as a regularizer, encouraging the model to learn from the limited available GT data while also exploring and generating plausible predictions for the unknown regions of the scene. 
The supervision is achieved by aligning the refined depth predictions with the sparse GT values in both depth latent space and pixel-wise depth through a sparse valid mask.
% The supervision is realized by aligning the refined depth with sparse GT in the depth latent space through a sparse valid mask. 
With the help of random crop, jitter, and flip augmentation in training, this process lets the generative model \textit{organize} the entire depth map instead of just regressing on known parts, which largely improves the visual quality of the depth prediction.

The proposed DiffusionDepth framework is evaluated on widely used public benchmarks KITTI~\cite{geiger2013kitti} and NYU-Depth-V2~\cite{redmon2016you}, covering both indoor and outdoor scenarios.
It could reach $0.298$ and $1.452$ RMSE on official offline test split respectively on NYU-Depth-V2 and KITTI datasets, which exceeds state-of-the-art (SOTA) performance. 
To better understand the effectiveness and properties of the diffusion-based approach for 3D perception tasks, we conduct a detailed ablation study. It discusses the impact of different components and design choices on introducing the diffusion approach to 3D perception, providing valuable insights as references for related tasks such as stereo and depth completion. 
The contribution of this paper could be summarized in threefold.

\begin{compactitem}
    \item This work proposes a novel approach to monocular depth estimation by reformulating it as an iterative diffusion-denoising problem with visual guidance.
    \item 
    %Experimental results suggest that, by refining the depth prediction in coarse and fine details, we achieve state-of-the-art performance on both online and offline evaluations, with affordable inference costs.
    Experimental results suggest DiffusionDepth achieves state-of-the-art performance on both offline and offline evaluations with affordable inference costs.
    \item This is the first work introducing the diffusion model into depth estimation, providing extensive ablation component analyses, and valuable insights for potentially related 3D vision tasks.
    %, such as stereo and depth completion.

\end{compactitem}

%-------------------------------------------------------------------------

\input{text/relatedworks}

\input{text/methodology}

\input{text/experiment}

\input{text/conclusion}

{\small
\bibliographystyle{ieee_fullname}
\bibliography{main.bib}}

\clearpage

\appendix
\input{text/appendix}

\end{document}

%% file: text/relatedworks.tex
\section{Related Works}

\textbf{Monocular Depth Estimation} is an important task in computer vision that aims to estimate the depth map of a scene from a single RGB image. % Over the years, various approaches have been proposed to tackle this challenging problem. 
Early approache~\cite{saxena2005learning} utilized Markov random field to predict depth, while more approaches~\cite{eigen2014depth,qi2018geonet,fu2018deep} leverage deep convolutional neural networks (CNNs) to achieve drastic performance.
One popular approach is to formulate monocular depth estimation as a dense per-pixel regression problem. Many methods, including RAP~\cite{zhang2019pattern}, DAV~\cite{huynh2020dav}, and BTS~\cite{lee2019bts}, have achieved impressive performance using this approach. Some follow-up approaches, such as UnetDepth~\cite{guo2018learning}, CANet~\cite{yan2021channel}, and BANet~\cite{aich2021bidirectional}, focus on modifying the backbone structure to enhance visual features.
Recently, transformer structures have been introduced in monocular depth estimation, where DPT~\cite{ranftl2021dpt}, and PixelFormer~\cite{agarwal2023attention} have shown improved performances.
To increase the robustness of monocular depth estimation, some methods introduce additional constraints such as uncertainty (UCRDepth~\cite{shao2023urcdc}) or piecewise planarity prior (P3Depth~\cite{patil2022p3depth}). NewCRFs~\cite{DBLP:journals/corr/abs-2203-01502} proposes window-separated Conditional Random Fields (CRF) to enhance the local space relation with neighboring pixels.
AdaBins~\cite{bhat2021adabins} and Binsformer~\cite{li2022binsformer} revisited ordinal regression networks and reformulate the task as a classification-regression task by calculating adaptive bins based on image content to estimate depth. 
VA-Depth~\cite{liu2023va} first introduces variational inference into refine depth prediction. 
We further introduce the diffusion approach to this task and leverage powerful generative capacity to generate highly-refined depth prediction. 

\textbf{Diffusion Model for Perception Tasks }
Although Diffusion models have achieved great success in image generation~\cite{ho2020denoising, song2021scorebased, dhariwal2021diffusion}, their potential for discriminative tasks remains largely unexplored. 
The improved diffusion process~\cite{song2020denoising} has made inference times to become more affordable for perception tasks, which has accelerated the exploration.
Some initial attempts have been made to adopt diffusion models for image segmentation tasks~\cite{wolleb2021diffusion, baranchuk2022labelefficient, graikos2022diffusion, kim2022diffusion, brempong2022denoising, amit2021segdiff, chen2022generalist}. 
These segmentation tasks are processed in an image-to-image style.
DiffusionDet~\cite{chen2022diffusiondet} first extends the diffusion process into generating detection box proposals. 
% Intead of adpoting diffusion process as a generative head, we formulate the denoising process as conditioned depth refinement. 
% However, none of the previous works have explored reformulating the diffusion into depth estimation tasks. 
We propose to use the diffusion model for denoising the input image as a conditioned depth refinement process, instead of adopting it as a normal generative head.
To the best of our knowledge, this is the first work introducing the diffusion model into monocular depth estimation.

%% file: text/methodology.tex
\section{Methodology}

% DiffusionDepth model, which takes a monocular visual input and uses a backbone module to extract multi-scale features. These features are then aggregated through an attention mechanism and a feature pyramid network (FPN) to construct visual guidance conditions. The diffusion head of the model iteratively refines the depth map by removing noise from an initial estimate and improving the accuracy of depth predictions, guided by the visual information available in the input image. This is achieved using a proposed monocular conditioned denoising block (MCDB) as shown in the figure. The MCDB utilizes the visual guidance conditions constructed by the Multi-Scale FPN to perform guided denoising, resulting in a refined depth prediction that is more accurate than the initial estimate. 

% The DiffusionDepth model is an innovative approach to monocular depth estimation that leverages the power of deep learning to produce highly accurate depth maps from a single input image.

\subsection{Task Reformulation}
\label{subsec:reformulation}

% \noindent \textbf{Object detection.}
% We reformulate monocular depth estimation as a denoising diffusion process.
% Given input image $\bm{c}$, the monocular depth estimation task is normally formulated as $p(\bm{x}|c)$, where $x$ is the desired 

% \textbf{Diffusion model.}
% Diffusion models~\cite{sohl2015deep, ho2020denoising, song2019generative, song2021denoising} are likelihood-based models inspired by nonequilibrium thermodynamics~\cite{song2019generative, song2020improved}. 

\paragraph{Preliminaries}
Diffusion models~\cite{sohl2015deep, ho2020denoising, song2019generative, song2021denoising} are a class of latent variable models. It is normally used for generative tasks, where neural networks are trained to denoise images blurred with Gaussian noise by learning to reverse the diffusion process.
The diffusion process $q(\bm{x}_t | \bm{x}_0)$ as defined in Eq.~\ref{eq:noise_process},
% Neural networks are trained to denoise images blurred with Gaussian noise by learning to reverse the diffusion process

\begin{equation}
\label{eq:noise_process}
    q(\bm{x}_t | \bm{x}_0) \coloneqq \mathcal{N}(\bm{x}_t | \sqrt{\bar{\alpha}_t} \bm{x}_0, (1 - \bar{\alpha}_t) \bm{I}),
\end{equation}
iteratively adds noise to desired image distribution $\bm{x}_0$ and get latent noisy sample $\bm{x}_t$ for $t\in\{0, 1, ...,T\}$ steps.
% adds noise for $t$ steps to desired image distribution $\bm{x}_0$
$\bar{\alpha}_t \coloneqq \prod_{s=0}^{t} \alpha_s = \prod_{s=0}^{t} (1 - \beta_s)$ and $\beta_s$ represents the noise variance schedule~\cite{ ho2020denoising}.
In the denoising process, neural network $\bm{\mu}_\theta(\bm{x}_t, t)$ is trained to reverse $\bm{x}_0$ by interactively predicting $\bm{x}_{t-1}$ as below.  
\begin{equation}
    p_{\theta}(\bm{x}_{t-1}|\bm{x}_t) \coloneqq \mathcal{N}(\bm{x}_{t-1};\bm{\mu}_{\theta}(\bm{x}_t, t), \bm{\sigma}_t^2 \bm{I}), \label{eq:ptheta}
\end{equation}
where $\bm{\sigma}_t^2$ denotes the transition variance. Sample $\bm{x}_0$ is reconstructed from prior noise $\bm{x}_T$ an mathematical inference process~\cite{ho2020denoising, song2021denoising} iteratively, \ie,  $\bm{x}_T \rightarrow \bm{x}_{T-\Delta} \rightarrow ... \rightarrow \bm{x}_0$. 
%by minimizing the training objective with $\ell_2$ loss~\cite{ho2020denoising}:
%\begin{equation}
%    \mathcal{L}_\text{train} =  \frac{1}{2}|| f_\theta(\bm{x}_t, t) - \bm{x}_0 ||^2.
%\end{equation}

% \cref{appendix:diffusion_formula}

% \noindent \textbf{Conditioned Denoising}\quad
\paragraph{Denoising as Depth Refinement}
Given input image $\bm{c}$, the monocular depth estimation task is normally formulated as $p(\bm{x}|\bm{c})$, where $\bm{x}$ is the desired depth map.
We reformulate the depth estimation as a visual-condition guided denoising process which refines the depth distribution $\bm{x}_t$ iteratively as defined in Eq.~\ref{eq:conditiondenoising} into the final depth map $\bm{x}_0$.
\begin{equation}
    p_{\theta}(\bm{x}_{t-1}|\bm{x}_t, \bm{c}) \coloneqq \mathcal{N}(\bm{x}_{t-1};\bm{\mu}_{\theta}(\bm{x}_t, t, \bm{c}), \bm{\sigma}_t^2 \bm{I}), \label{eq:conditiondenoising}
\end{equation}
where model $\bm{\mu}_\theta(\bm{x}_t, t, \bm{c})$ is trained to refine depth latent $\bm{x}_t$ to $\bm{x}_{t-1}$. 
To accelerate the denoising process, we utilized the improved inference process from DDIM~\cite{song2020improved}, where it set $\bm{\sigma}_t^2 \bm{I}$ as 0 to make the prediction output deterministic. 

% from noisy boxes $\bm{x}_t$, conditioned on the corresponding image $\bm{x}$. The corresponding category label $\bm{c}$ is produced accordingly.

\subsection{Network Architecture}
\label{subsec:networkarch}
We use Swin Transformer~\cite{liu2021swin} as shown in Fig.~\ref{fig:mainstructure} as an example to illustrate the feature extraction. 
The input image is patched and projected into visual tokens with position embedding.  
The backbone extracts visual features at a different scale to maintain coarse and fine details of the input scene. 
Based on extracted multi-scale features, we employ hierarchical aggregation and heterogeneous interaction (HAHI~\cite{li2022depthformer}) to enhance features between scales. Feature pyramid neck~\cite{lin2017feature} is applied to aggregate features into monocular visual condition. 
The \textbf{visual condition} is the aggregated feature map with a shape $\frac{H}{4}\times \frac{W}{4} \times c$, where $H, W$ are respectively the height and width of the monocular image input, and $c$ is the channel dimension the feature. 
The proposed DiffusionDepth model is suitable for most visual backbones which could extract multi-scale features. 
According to extensive experiments, other backbones such as ResNet~\cite{he2016deep}, Efficient~\cite{tan2019efficientnet}, and ViT~\cite{dosovitskiy2020vit} could achieve competitive performance as well. 

\subsection{Monocular Conditioned Denoising Block}

\begin{figure}[hbpt]
\begin{center}
% \fbox{\rule{0pt}{1.5in} \rule{0.9\linewidth}{0pt}}
   \includegraphics[width=0.5\textwidth]{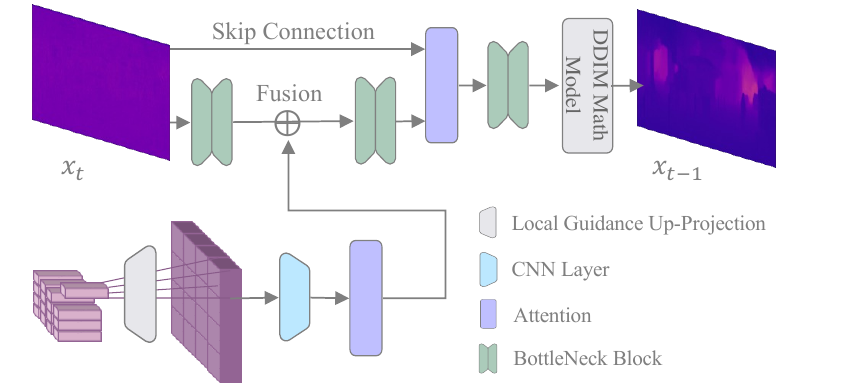}
\end{center}
   \caption{Illustration of Monocular Conditioned Denoising Block. Visual condition is fused with depth latent through hierarchically.}
\label{fig:mcdb}
\end{figure}

As mentioned above (Section~\ref{subsec:reformulation}), we formulate the depth estimation as a denoising process $p_{\theta}(\bm{x}_{t-1}|\bm{x}_t, \bm{c})$, which iteratively refines depth latent $\bm{x_t}$ and improves the prediction accuracy, guided by the visual information available in the input image. Specifically, it is achieved by neural network model $\bm{\mu}_{\theta}(\bm{x}_t, t, \bm{c})$ which takes visual condition $\bm{c}$ and current depth latent $\bm{x}_t$ and predict the distribution $\bm{x}_{t-1}$. 
The monocular visual condition $\bm{c} \in \mathbb{R}^{\frac{H}{4}\times \frac{W}{4} \times c}$ is constructed through multi-scale visual feature aggregation (Section~\ref{subsec:networkarch}).
We introduce Monocular Conditioned Denoising Block (MCDB) as shown in Fig.~\ref{fig:mcdb} to achieve this process. 

%-------------------------------------------------------------------------
%-------------------------------------------------------------------------

Since the depth prediction task normally requires low inference time for practical utilization, we design the denoising head in a \textbf{light-weighted} formation. 
The visual condition $\bm{c}$ is actually aggregated feature map with a lower resolution which has a strong local relation to the depth latent $\bm{x}_t$ to be denoised. We first use a local projection layer to upsample the condition $c$ into the same shape with the depth latent $\bm{x}_t \in \mathbb{R}^{\frac{H}{2}\times \frac{W}{2} \times d}$ while maintaining the local relation between features.
The projected condition is directly fused with the depth latent $\bm{x}_t$ by performing element-wise summation through a CNN block and a self-attention layer. 
The fused depth latent is processed by a normal BottleNeck~\cite{he2016deep} CNN layer and channel-wise attention with the residual connection. 
The denoising output $\bm{x}_{t-1}$ is calculated by applying DDIM~\cite{song2020improved} inference process according to prefixed diffusion schedule $\beta, \alpha$ on model outputs.

\subsection{Diffusion-Denosing Process}
The diffusion process $q(\bm{x}_t | \bm{x}_0)$ and denoising process $ p_{\theta}(\bm{x}_{t-1}|\bm{x}_t, \bm{c})$ are respectively defined in Eq.~\ref{eq:noise_process}, Eq.~\ref{eq:conditiondenoising}. 
Trainable parameters are mainly the conditioned denoising model $\bm{\mu}_\theta(\bm{x}_t, t, \bm{c})$ and visual feature extractors defined above. 
The model is trained by minimizing the $L_D$ loss between diffusion results and denoising prediction in Eq.~\ref{eq:loss-ddim}. 
\begin{equation}
L_{\mathrm{ddim}} = \left\Vert\boldsymbol{x}_{t-1} - \bm{\mu}_{\theta}(\boldsymbol{x}_t, t, \bm{c})\right\Vert^2\label{eq:loss-ddim}
\end{equation}
where diffusion result $\boldsymbol{x}_{t-1}$ could be calcuated through diffusion process defined in Eq.~\ref{eq:noise_process} by sampling set of $t$. 
It actually supervises the depth of the latent at each step after refinement by reversing the diffusion process. 

\paragraph{Depth Latent Space} Many of the previous constraint-based or classification-based methods are not good at generating depth maps in high resolution.
We employ a similar structure with latent diffusion~\cite{rombach2022high}, where both diffusion processes $q(\bm{x}_t | \bm{x}_0)$ and denoising process $ p_{\theta}(\bm{x}_{t-1}|\bm{x}_t, \bm{c})$ are performed in encoded latent depth space. 
The refined depth latent $\bm{x}_0 \in \mathbb{R}^{\frac{H}{2}\times \frac{W}{2} \times d}$ with latent dimension $d$ is transferred to depth estimation $\bm{de} \in \mathbb{R}^{{H}\times {W} \times 1}$ through a depth decoder. 
The depth decoder is composed of sequentially connected 1x1 convolution, 3x3 de-convolution, 3x3 convolution, and a Sigmoid~\cite{nwankpa2018activation} activation function. The depth is calculated through Eq.~\ref{eq:output}. 
\begin{equation}\label{eq:output}
    \bm{de} = 1/ {\mathrm{sig}(\bm{x}_0)}.\mathrm{clamp(\eta)} -1,
\end{equation}
where $\eta$ is the max output range. We set $\eta = 1e^6$ for both indoor and outdoor scenarios.
Considering the sparsity in GT depth $\hat{\bm{de}}$, we use a BottleNeck CNN block with channel dimension $d$ and kernel size $1\times1$ to encode the depth GT into depth latent $\hat{\bm{x}}_0$. 
The decoder and encoder are trained directly in end-to-end formation by minimizing the direct pixel-wise depth loss defined in Eq.~\ref{eq:pixel}. 
\begin{equation}\label{eq:pixel}
  L_{\mathrm{pixel}} = \sqrt{\frac{1}{T}\sum_i \bm{\delta}_i^2 + \frac{\lambda}{T^2} (\sum_i  \bm{\delta}_i)^2},
\end{equation}
where $\bm{\delta}_i=\hat{\bm{de}} - \bm{de}$ is the pixel-wise depth error on valid pixels, $\lambda$ is set to $0.85$~\cite{li2022depthformer} for all experiments. $T$ is the total number of the valid pixels. 
The supervision is also applied to both latent spaces through L2 loss between encoded GT latent $\hat{\bm{x}}_0$ and depth latent ${\bm{x}}_0$ through a valid mask as defined in Eq.~\ref{eq:latent}.
\begin{equation}\label{eq:latent}
   L_{\mathrm{latent}} = \left\Vert\bm{x}_{0} - \hat{\bm{x}_0}\right\Vert^2
\end{equation}
\begin{equation}\label{eq:totalloss}
    L = \lambda_1 L_{ddim} + \lambda_2 L_{pixel} + \lambda_3 L_{pixel},
\end{equation}
The DiffusionDepth is trained by combining losses through a weighted sum and minimizing the $L$ defined in Eq.~\ref{eq:totalloss}. 

\paragraph{Self-Diffusion} One severe problem of adopting generative methods into depth prediction is the \textbf{sparse ground truth} (GT) depth value problem, which is prevalent in outdoor scenarios where only a fraction of pixels have GT depth values (typically around $3.75-5\%$ in datasets such as KITTI depth). This sparsity can lead to mode collapse during normal generative training.
To tackle this issue, DiffusionDepth introduces a self-diffusion process. 
Rather than directly diffusing on the encoded sparse GT depth in latent space, the model gradually adds noise to the refined depth latent $\bm{x}_0$ from the current denoising output. 
With the help of random crop, jitter, and flip augmentation in training, this process allows the model ``organize" the entire depth map instead of just regressing on known parts, which largely improves the visual quality of the depth prediction.
According to our experiments, for indoor sceneries with dense GT values, diffusion on either refined depth or GT depth is feasible.

%% file: text/experiment.tex
\begin{figure*}
\includegraphics[width=1\textwidth]{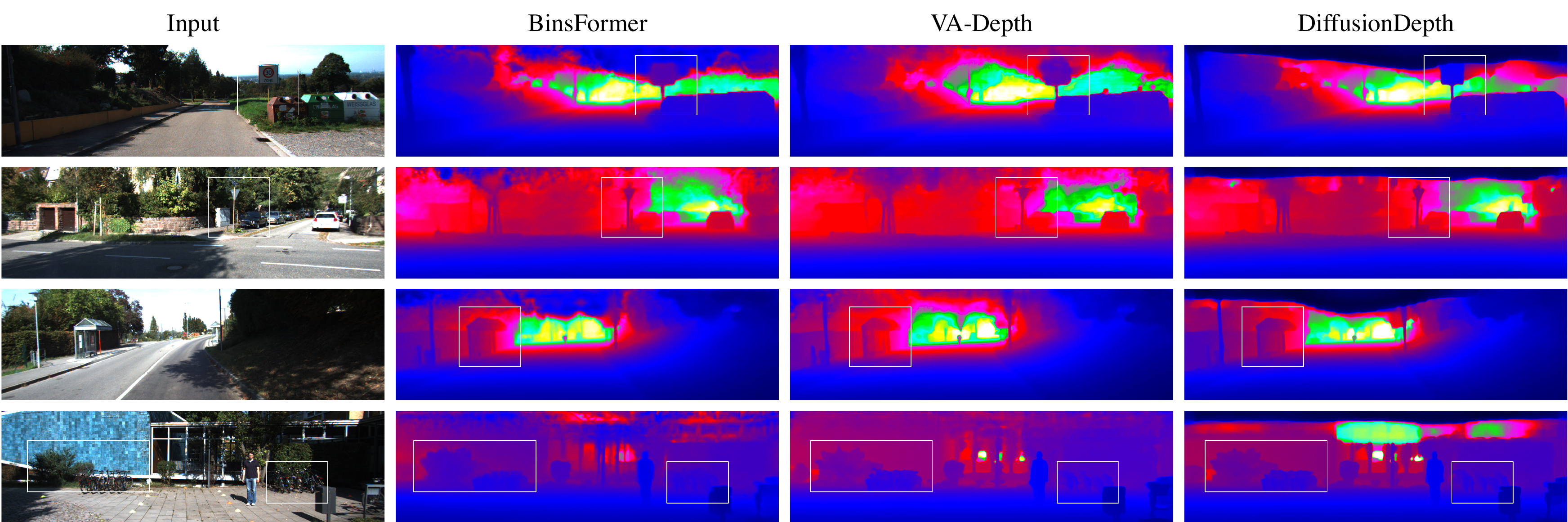}
\caption{Qualitative comparison of proposed DiffusionDepth on the KITTI outdoor driving scenarios against two representative methods, BinsFormer (classification-regression based) and VA-Depth (Variational Refine). We highlight the details with white boxes. The visualization is from the best online results for a fair comparison.\label{fig:visualizationkitti}}
\end{figure*}

\input{tables/offlinekitti}

\section{Experiment}
\label{sec:exepriment}

% \subsection{Experimental Setup}

\paragraph{Dataset} We conduct detailed experiments on outdoor and indoor scenarios to report an overall evaluation of the proposed DiffusionDepth and its properties.

\textbf{KITTI dataset} is captured from outdoor with driving vehicles~\cite{geiger2013kitti} with depth range 0-100m. The image resolution is around 1216 × 352 pixels with sparse GT depth (density $3.75\%$ to $5\%$). We evaluate on both Eigen split~\cite{eigen2014depth} with 23488 training image pairs and 697 testing images and official split~\cite{geiger2013kitti} with 42949 training image pairs, 1000 validation images, and 500 testing images. 

\textbf{NYU-Depth-v2 dataset} is collected from indoor scenes at a resolution of 640 × 480 pixels~\cite{Silberman:ECCV12} and dense depth GT (density $>95\%$). Following prior works, we adopt the official split and the dataset processed by Lee \etal ~\cite{lee2021patch}, which contains 24231 training images and 654 testing images.

\paragraph{Implementation details}

DiffusionDepth is implemented with the Pytorch~\cite{paszke2019pytorch} framework. 
% The code is accessible through an anonymous link~\footnote{\href{https://anonymous.4open.science/r/DiffusionDepth-4951}{https://anonymous.4open.science/r/DiffusionDepth-4951}}. 
The \textbf{codes} are provided in supplementary materials and will be \textbf{open-sourced} after the review process to contribute to the community. 
We train the entire model with batch size 16 for 30 epochs iterations on a single node with 8 NVIDIA A100 40G GPUs. 
We utilize the AdamW optimizer~\cite{kingma2014adam} with ($\beta_1$, $\beta_2$, $w$) = (0.9, 0.999, 0.01), where $w$ is the weight decay. 
The linear learning rate warm-up strategy is applied for the first $15\%$ iterations. The cosine annealing learning rate strategy is adopted for the learning rate decay from the initial learning rate of $1e-4$ to $1e-8$. 
We use L1 and L2 pixel-wise depth loss at the first $50\%$ training iterations as auxiliary subversion. 
% For the NYU-Depth-v2 dataset, we utilize the official 25 classes divided by folder names for the auxiliary scene understanding task. For KITTI, since the outdoor dataset is tough to classify, we omit the scene classification loss and only use ground truth depth to provide supervision.
For the KITTI dataset, we sequentially utilize the random crop with size $706\times352$, color jitter with various lightness saturation, random scale from $1.0$ to $1.5$ times, and random flip for training data augmentation. 
For the NYU-Depth-V2 dataset, we use the same augmentation with the random crop with size $512\times340$. 

\textbf{Visual Condition:} DiffusionDepth is compatible with any backbone which could extract multi-scale features. 
Here, we respectively evaluate our model on the standard convolution-based ResNet~\cite{he2016deep} backbones and transformer-based Swin~\cite{liu2021swin} backbones. 
We employ hierarchical aggregation and heterogeneous interaction (HAHI~\cite{li2022depthformer}) neck to enhance features between scales and feature pyramid neck~\cite{lin2017feature} to aggregate features into monocular visual condition. 
The visual condition dimension is equal to the last layer of the neck. 
We respectively use channel dimensions $[64,128,256,512]$ and $[192, 384, 768, 1536]$
for ResNet and Transformer backbones. 

\textbf{Diffusiong Head:} We use the improved sampling process~\cite{song2020improved} with 1000 diffusion steps for training and 20 inference steps for inference. 
The learning rate of the diffusion head is 10 times larger than the backbone parameters. 
The dimension $d$ of the encoded depth latent is 16 with shape $\frac{H}{2},\frac{W}{2}, d$, we conduct detailed ablation to illustrate different inference settings. The max depth value of the decoder is $1e6$ for all experiments. 

\subsection{Benchmark Comparison with SOTA Methods}

\paragraph{Evaluation on KITTI Dataset}
We first illustrate the efficiency of DiffusionDepth by comparing it with previous state-of-the-art (SOTA) models on KITTI offline Eigen split~\cite{eigen2014depth} with an evaluation range of 0-80m and report results in Tab.~\ref{tb:offlinekitti}. 
It is observed that DiffusionDepth respectively reaches $0.050$ absolute error and $2.016$ RMSE on the evaluation, which exceeds the current SOTA results URCDC-Depth (RSME $2.032$) and VA-Depth (RSME $2.090$). 
On official offline split~\cite{geiger2013kitti} in Tab.~\ref{tb:offlinekitti} with evaluation range 0-50m,
our proposal reaches $0.041$ absolute related error and $1.452$ RMSE on the evaluation, which largely outperforms the current best URCDC-Depth ($0.049$ rel and $1.528$ RMSE) by a large margin.
%($\Delta 16.32\%$ and $\Delta 4.9\%$). 
This suggests that DiffusionDepth has even better performance in estimating depth with a closer depth range which is valuable for practical usage. 
This property is rational since the diffusion approach brings a stronger generative ability to the task. 
\input{tables/onlinekitti}

%We also conduct a comparison by submitting results on official servers to perform \textbf{KITTI Online Evaluation} with 500 unseen images. 
%The reports are shown in Tab.~\ref{tb:onlinekitti}. However, in these 500 images, the proposed model is slightly lower than VA-depth and URCDC-depth.   
%Since this paper is to propose the diffusion approach to the depth estimation task with only naive aggregated visual features as guidance without complicated long-range attention or constraint priors. 
%It is noted that the proposed diffusion head is compatible with these SOTA depth feature extraction methods. 
%The performance of the diffusion approach could be further boosted by these more advanced visual conditions from other papers, such as using bins~\cite{li2022binsformer} or uncertainty~\cite{shao2023urcdc} as the denoising guidance. 

Online evaluation is conducted by submitting results to the official servers for \textbf{KITTI Online evaluation} on 500 unseen images. 
The results are shown in Tab.\ref{tb:onlinekitti}, where the proposed model slightly underperformed compared to VA-depth and URCDC-depth. 
However, it's important to note that our approach only uses aggregated visual features as guidance and doesn't incorporate complicated long-range attention or constraint priors like these SOTA methods. As our diffusion head is compatible with these advanced depth feature extraction techniques, incorporating them could further improve the performance of our approach.

\noindent\textbf{Qualitative Comparison on KITTI Datset} is reported in Fig.~\ref{fig:visualizationkitti}. Here, we show the improved visual quality brought by the diffusion-denoising process. 
The clarity of the objects regarding both the edges and the shape has been significantly improved. 
For example, on the first row, both BinsFormer and VA-Depth have significant blur on the signpost. Diffusion depth predicts a sharp and accurate shape for it. 
Classification-based methods are suffered from visible noise in the depth map. 
As we mentioned above, one significant advantage of introducing the diffusion-denoising approach is that we could acquire a highly-detailed depth map with good visual quality and clear shapes for practical utilization. The proposed diffusion head could also be combined with other methods, such as bins to improve the visual quality.

\begin{figure}[hbpt]
    \includegraphics[width=1\columnwidth]{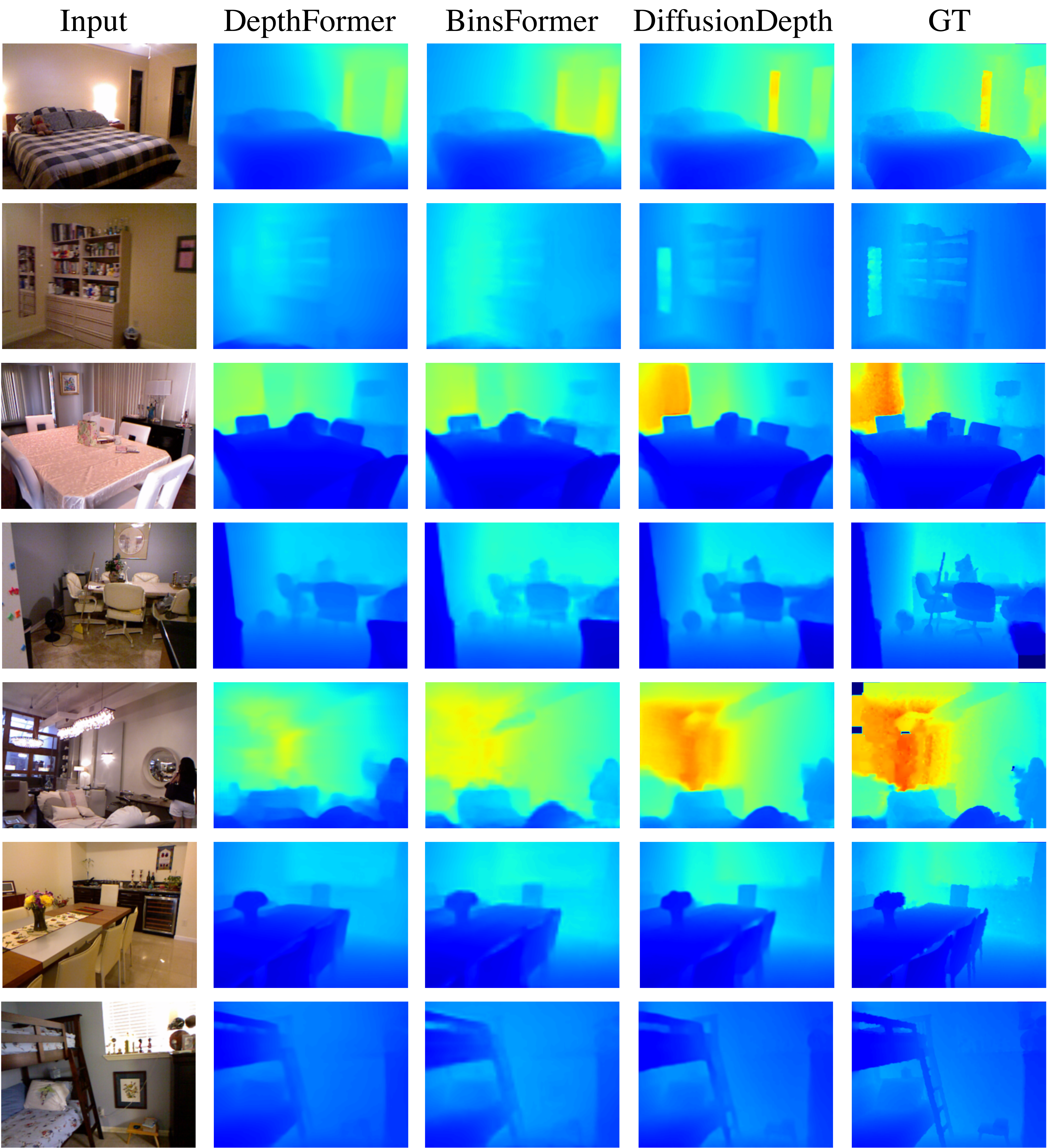}
    \caption{Qualitative depth results on the NYU-Depth-v2 dataset.\label{fig:visualnyu}} 
\end{figure}

\input{tables/offlinenyudepth_v2}

%\noindent\textbf{Evaluation on NYU-Depth-V2 Dataset} 
\paragraph{Evaluation on NYU-Depth-V2 Dataset} 
We evaluate the proposed DiffusionDepth on the NYU-Depth-v2 dataset~\cite{silberman2012indoor} to demonstrate the effectiveness of our proposal. 
The results are reported in Table~\ref{fig:visualnyu}. 
It suggests that the diffusion-denoising approach has even higher improvement than outdoor scenarios, where it respectively achieves $0.85$ absolute related error and $0.295$ RSME score which exceeds the previous SOTA
We think this phenomenon is rational since indoor scenarios mostly have dense depth GT values, which is naturally suitable for generative models. 
It is noted that for datasets with dense GTs, direct diffusion on GT value is also feasible with comparable results. 
% Here, for better consistency, we still report the results for self-diffusion. 
To give a more direct illustration of the proposed DiffusionDepth.  \textbf{We display qualitative depth comparisons in Fig.~\ref{fig:visualnyu}}. 

% \subsection{Diffusion Properties}
\subsection{Ablation Study}

\paragraph{Qualitative Study of Denoising Process} To give an intuitive understanding of how the denoising process refines the depth prediction step by step, we visualize the denoising process in Fig.~\ref{fig:diffprocess}. 
It shows that the process first initializes ($t<10$) the shapes and edges from random depth distribution. 
Then the guided denoising model refines the depth values and corrects distance relations step by step.
This process is more like first recognizing the shape of the desired scenery and then considering the depth relations between these objects with visual clues. 
The learning process is impressive. 
One interesting problem is that the denoising process is even faster in more complicated outdoor scenarios (KITTI). Although the mediate results are slightly lower, the denoising steps larger than 15 could achieve competitive results on the KITTI dataset. 

\iffalse
\begin{figure}[hbpt]
    \includegraphics[width=1\columnwidth]{img/diffusionprocess.pdf}
    \caption{Visualization of the diffusion process.\label{fig:diffprocess}} 
\end{figure}
\begin{figure}[hbpt]
    \includegraphics[width=0.85\columnwidth]{img/diffusionprocessnyu.pdf}
    \caption{Visualization of the diffusion process.\label{fig:diffprocess}} 
\end{figure}
\fi

\begin{figure}[t]
  \centering
 \subfigure[Visualization on KITTI Dataset]{
\includegraphics[width=1\columnwidth]{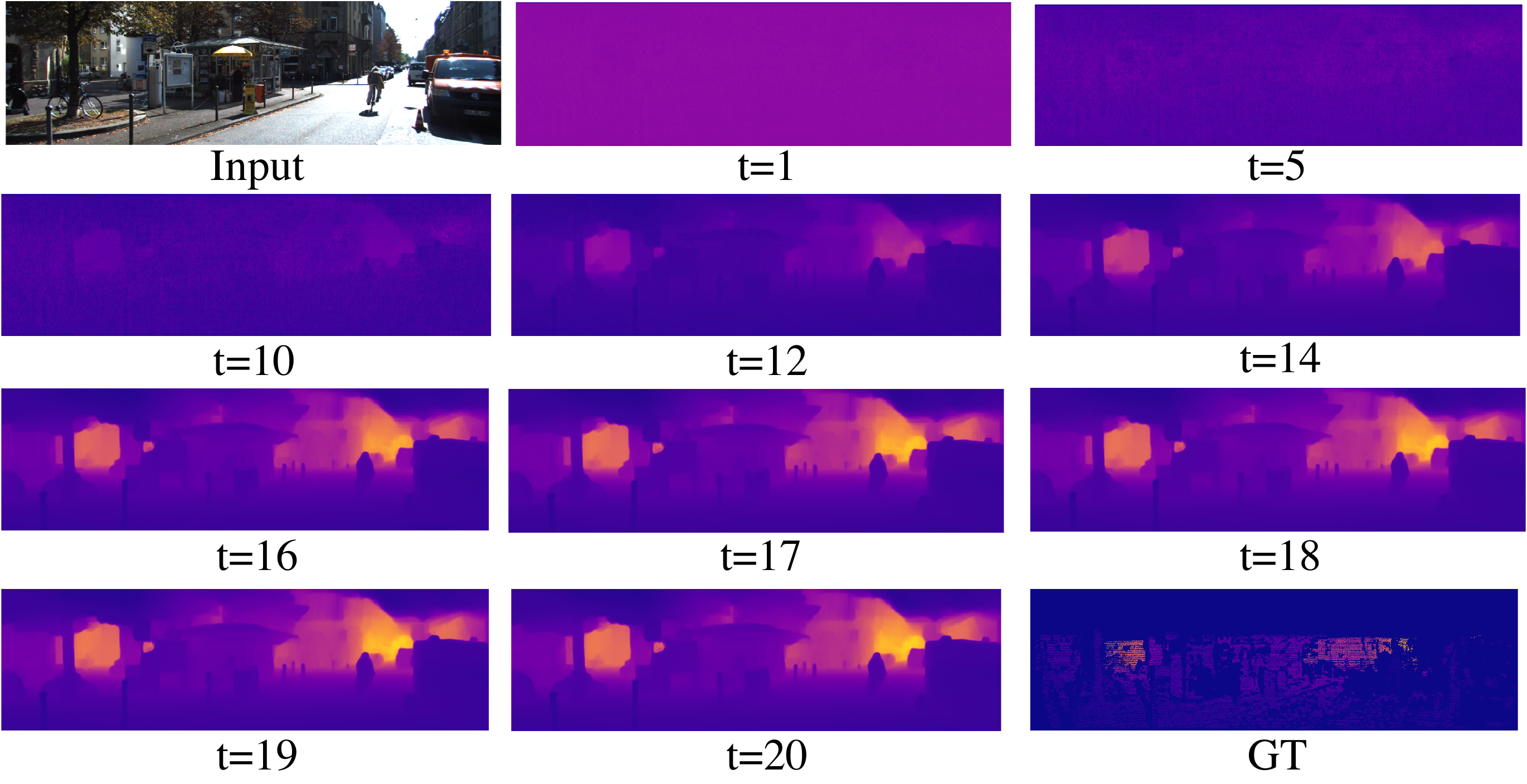}\label{fig:kittidiff}
}
\hfill
 \subfigure[Visualization on NYU-Depth-V2 Dataset]{
\includegraphics[width=1\columnwidth]{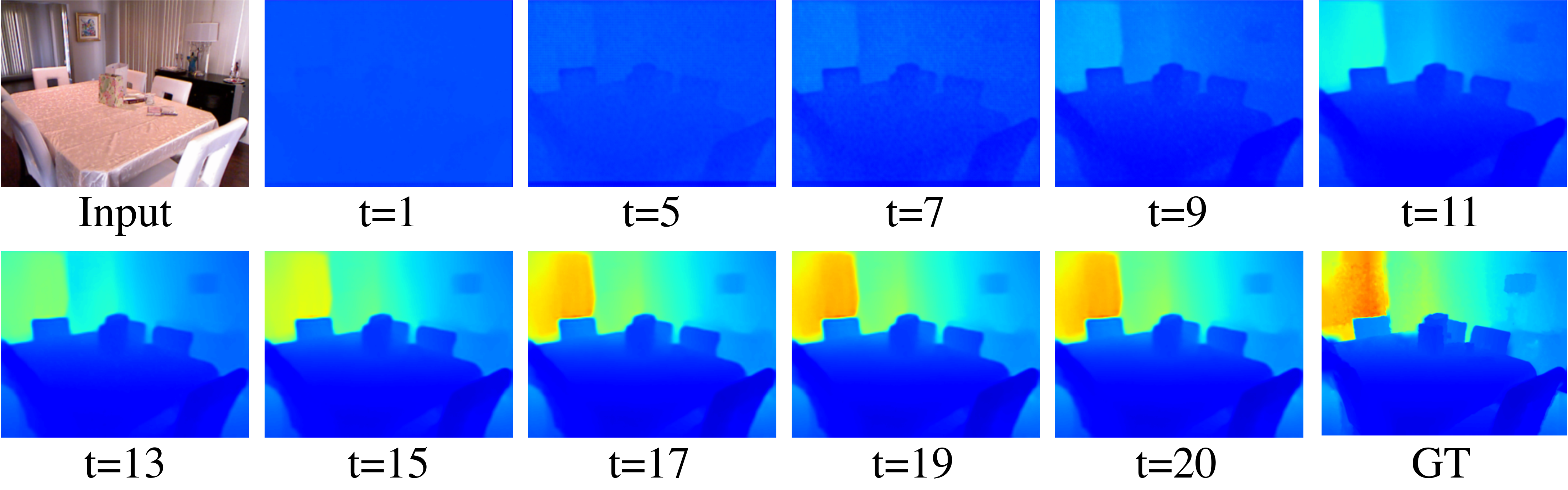}\label{fig:nyudiff}
}
  \caption{Visualization of the denoising process with 20 inference steps, where $t$ denotes the current step. It gives an intuitive illustration of how the depth estimation is refined iteratively. \label{fig:diffprocess}}
\end{figure}

\noindent\textbf{Denoising Inference} 
To further reveal the properties of using different inference steps, we conduct an ablation study on different inference settings. Lower inference steps could benefit the practical usage with lower GPU memory consumption and faster inference speed. 
We consider two settings, 1) train with 1000 diffusion steps and 20 inference steps and change the inference step, 2) train with different inference steps. 
The ablation is conducted on the NYU-Depth-V2 dataset, where the variations of the metrics are reported in Tab.~\ref{tb:disproperties}.
We fix diffusion to 1000 steps throughout the training. 
Directly changing the inference steps will lead to a severe performance drop. This observation is different from the diffusion approach on detection boxes~\cite{chen2022diffusiondet}, which could change inference steps once the model is trained. 
We think this observation is rational since directly denoising on the highly detailed depth map is closer to a generative task, rather than denoising on anchor boxes. 
However, we prove the feasibility of accelerating the inference by directly training the model with the desired inference setting, which only shows a slight performance drop. 
% However, we prove the feasibility of accelerating the training with different inference steps.

\begin{figure}[t]
\centering
    \includegraphics[width=0.95\columnwidth]{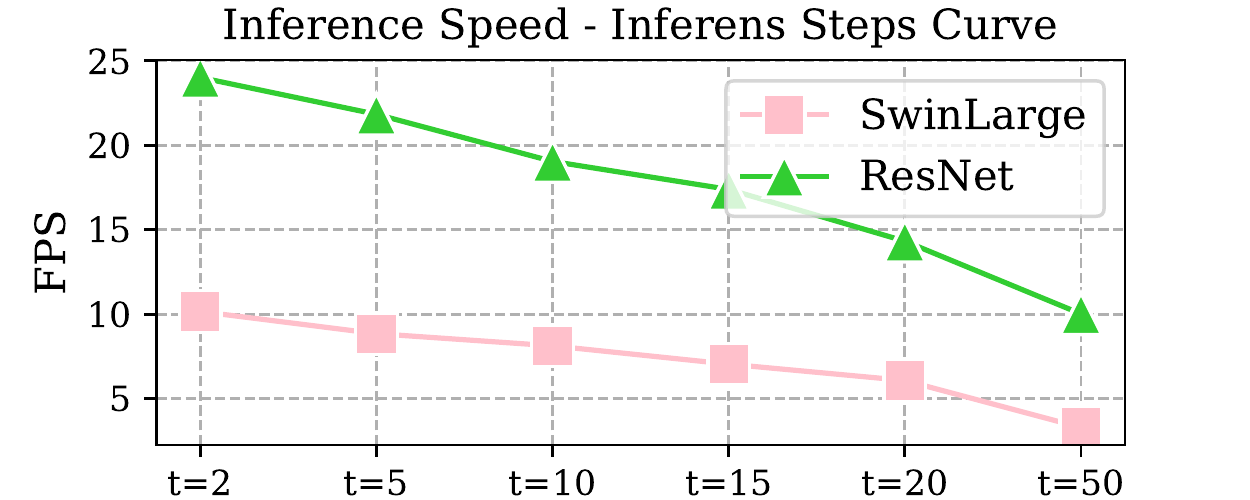}
    \caption{Inference speed with RTX 3090 GPU on KITTI dataset. \label{fig:inference}} 
\end{figure}

\paragraph{Inference Speed} 
Although the inference speed is one shortage of diffusion-based models, as shown by Fig.~\ref{fig:inference}, DiffusionDepth could reach 14 FPS and 5 FPS (Frame Per Second) respectively on ResNet Backbones and Swin Backbones with 20 inference steps, which is feasible for practical usage.
With acceleration, the speed could be faster. 

\hfill
\input{tables/diffusionproperty}

\input{tables/diffusionprocess}
% \paragraph{Diffusion}
\noindent\textbf{Diffusion}
As we mentioned, we employ a self-diffusion formation to add noise on refined depth latent rather than directly on the sparse depth.
In outdoor scenarios, sparse depth GT will lead to severe mode collapse. 
We illustrate the phenomenon by comparing different diffusion methods on both KITTI and NYU-Depth-V2 datasets.
The results are reported in Tab.~\ref{tb:diffusionmethod}, where diffusing on sparse GT on the KITTI dataset is not even converging. 
Both diffusion approaches could achieve competitive results on the NYU-Depth-V2 dataset with dense GT depth values. 

\input{tables/ablationlatent}

\noindent\textbf{Depth Latent Space}
We report the impact of different depth encoder-decoders with different down-sampling rates and report the results in Tab.~\ref{tb:ablationstructure}. It suggests that both $\times 4 $ and $\times 2$ encoder-decoder pairs could achieve competitive results. However, depth latent space with higher resolution slightly outperforms others. 

\noindent\textbf{Compatibility with other Visual Conditions} We report the compatibility of the proposed model in Tab.~\ref{tb:ablationstructure}.
It suggests that the proposed DiffusionDepth head could achieve good results on both CNN-based models (Res34, Res50) and transformer-based models (Swin). 
We also conduct simple experiments to illustrate our proposal could utilize visual depth conditions such as Bins~\cite{li2022binsformer} as the denoising guidance. It could also achieve competitive performance. 

Please refer to Appendix for more detailed hyper-parameters on different hyper-parameters and more visualization results of DiffusionDepth.

%% file: tables/offlinekitti.tex
\begin{table*}[hbtp]
\caption{Evaluation metrics on the offline KITTI dataset, Eigen split~\cite{eigen2014depth} and official offline split~\cite{geiger2013kitti}. The metrics of comparison metrics come from corresponding original papers. “-” indicates not applicable. The best results are highlighted in bold. \label{tb:offlinekitti} }
\centering
\begin{tabular}{c|c|cccc|ccc}
\Xhline{1.2pt}
\textbf{Method} & \textbf{Cap} & \textbf{Abs Rel $\downarrow$} & \textbf{Sq Rel $\downarrow$} & \textbf{RMSE $\downarrow$} & \textbf{RMSE log $\downarrow$} & \textbf{$\delta^1$ $\uparrow$} & \textbf{$\delta^2$ $\uparrow$} & \textbf{$\delta^3$ $\uparrow$} \\ \hline
\multicolumn{9}{c}{Eigen Split ~\cite{eigen2014depth}, evaluation range 0-80m }
\\
\hline
Eigen~\textit{et al.}~\cite{eigen2015predicting} & 0-80m & 0.203 & 1.548 & 6.307 & 0.282 & 0.702 & 0.898 & 0.967 \\
DORN~\cite{fu2018deep} & 0-80m & 0.072 & 0.307 & 2.727 & 0.120 & 0.932 & 0.984 & 0.994 \\
VNL~\cite{yin2019enforcing} & 0-80m & 0.072 & - & 3.258 & 0.117 & 0.938 & 0.990 & 0.998 \\
BTS~\cite{lee2019big} & 0-80m & 0.061 & 0.261 & 2.834 & 0.099 & 0.954 & 0.992 & 0.998 \\
PWA~\cite{lee2021patch} & 0-80m & 0.060 & 0.221 & 2.604 & 0.093 & 0.958 & 0.994 & \textbf{0.999} \\
TransDepth~\cite{yang2021transformer} & 0-80m & 0.064 & 0.252 & 2.755 & 0.098 & 0.956 & 0.994 & \textbf{0.999} \\
Adabins~\cite{bhat2021adabins} & 0-80m & 0.058 & 0.190 & 2.360 & 0.088 & 0.964 & 0.995 & \textbf{0.999} \\
P3Depth~\cite{patil2022p3depth} & 0-80m & 0.071 & 0.270 & 2.842 & 0.103 & 0.953 & 0.993 & 0.998 \\
DepthFormer~\cite{li2022depthformer} & 0-80m & 0.052 & 0.158 & 2.143 & 0.079 & 0.975 & {0.997} & \textbf{0.999} \\
NeWCRFs~\cite{Yuan_2022_CVPR} & 0-80m & 0.052 & 0.155 & 2.129 & 0.079 & 0.974 & {0.997} & \textbf{0.999} \\
PixelFormer~\cite{agarwal2023attention} & 0-80m & 0.051 & 0.149 & 2.081 & 0.077 & 0.976 & 0.997 & \textbf{0.999} \\
BinsFormer~\cite{li2022binsformer} & 0-80m & 0.052 & 0.151 & 2.098 & 0.079 & 0.974 & 0.997 & \textbf{0.999} \\
VA-Depth~\cite{liu2023va} & 0-80m & 0.050 & - & 2.090 & 0.079 & \textbf{0.977} & 0.997 & - \\
URCDC-Depth~\cite{shao2023urcdc} & 0-80m & {0.050} & {0.142} & {2.032} & {0.076} & \textbf{0.977} & {0.997} & \textbf{0.999} \\ \hline
\textbf{DiffusionDepth} (ours) & 0-80m & \textbf{0.050} & \textbf{0.141} & \textbf{2.016} & \textbf{0.074} & \textbf{0.977} & \textbf{0.998} & \textbf{0.999} \\ 
\Xhline{0.8pt}
\multicolumn{9}{c}{Official Offline Split ~\cite{geiger2013kitti}, evaluation range 0-50m }
\\
\hline
BTS~\cite{lee2019big} & 0-50m & 0.058 & 0.183 & 1.995 & 0.090 & 0.962 & 0.994 & 0.999 \\
PWA~\cite{lee2021patch} & 0-50m & 0.057 & 0.161 & 1.872 & 0.087 & 0.965 & 0.995 & 0.999 \\
TransDepth~\cite{yang2021transformer} & 0-50m & 0.061 & 0.185 & 1.992 & 0.091 & 0.963 & 0.995 & 0.999 \\
P3Depth~\cite{patil2022p3depth} & 0-50m & 0.055 & 0.130 & 1.651 & 0.081 & 0.974 & 0.997 & 0.999 \\
{URCDC-Depth}~\cite{shao2023urcdc} & 0-50m & {0.049} & {0.108} & {1.528} & {0.072} & {0.981} & {0.998} & \textbf{1.000} \\ \hline
\textbf{DiffusionDepth} (ours) & 0-50m & \textbf{0.041} & \textbf{0.103} & \textbf{1.418} & \textbf{0.069} & \textbf{0.986} & \textbf{0.999} & \textbf{1.000} \\
\Xhline{1.2pt}
\end{tabular}
\end{table*}

%% file: tables/onlinekitti.tex
\begin{table}[thbp]
 %The best results are highlighted in \textbf{bold}.}
	%\begin{center}
 \caption{Quantitative depth comparison on the official \textbf{online server of the KITTI dataset}.\label{tb:onlinekitti}}
		\smallskip
            \setlength\tabcolsep{2pt}
		\renewcommand{\arraystretch}{1.3}
		\resizebox{1\columnwidth}{!}{\begin{tabular}{c| c c c c  }	
				\Xhline{1.2pt}
				\textbf{Method} & \textbf{SILog} $\downarrow$ & \textbf{sqErrRel} $\downarrow$ & \textbf{absErrRel} $\downarrow$ & \textbf{iRMSE}$\downarrow$\\
				\hline						

				BTS~\cite{lee2019big}&11.67&9.04&2.21&12.23
				\\
				BANet~\cite{aich2021bidirectional}&11.61&9.38&2.29&12.23
				\\
				PackNet-SAN~\cite{guizilini2021sparse}&11.54&9.12&2.35&12.38
				\\
				PWA~\cite{lee2021patch}&11.45&9.05&2.30&12.32
				\\
				NeWCRFs~\cite{Yuan_2022_CVPR}&10.39&8.37&1.83&11.03
				\\
                  
                    PixelFormer~\cite{li2022binsformer}& 10.28 & 8.16 & 1.82 & 10.84	
				\\
                    BinsFormer~\cite{li2022binsformer}& 10.14	 & 8.23 & 1.69 & 10.90
				\\
                    P3Depth~\cite{patil2022p3depth}&12.82&9.92 & 2.53 & 13.71	
				\\
				{URCDC-Depth~\cite{shao2023urcdc}} &{10.03} &{8.24}&{1.74}&{10.71}
				\\
    {VA-Depth~\cite{liu2023va} } &\textbf{9.84} &\textbf{7.96}&{1.66}&\textbf{10.44}\\
    \hline
    \textbf{DiffusionDepth (ours)} &{9.85} &{8.06}&\textbf{1.64}&{10.58}\\
				\Xhline{1.2pt}		
		\end{tabular}}
	%\end{center}
	
\end{table}

%% file: tables/offlinenyudepth_v2.tex
\iffalse
\begin{table}[hbpt]
	\begin{center}
		%\smallskip
		\renewcommand{\arraystretch}{1.3}
		\resizebox{1.0\columnwidth}{!}{\begin{tabular}{c c c c c c c c c}
				\Xhline{1.2pt}
				ID & CD & UP & CU & CF& Abs Rel $\downarrow$ & RMSE $\downarrow$ &  ${\textbf{\rm{log}}_{\bm{{10}}}}$ $\downarrow$ & $\delta  < 1.25$ $\uparrow$\\
				\hline
				\hline				
				1&&& && 0.095 & 0.334 & 0.041 & 0.922\\
				% 2 &\cmark&&&& 0.063 &[0.061, 0.065]& 0.487 &[0.456, 0.518]& 5.183 &[4.977, 5.390]& 0.086 &[0.083, 0.089]& 0.967&[0.964, 0.971]\\	
				2 &\cmark&&&& 0.095 & 0.329 & 0.040 & 0.923\\	
				% 3 &\cmark&\cmark&&& 0.058 &[0.056, 0.060] & 0.439 &[0.409, 0.468]& 4.951 &[4.743, 5.159]& 0.081 &[0.078, 0.084]& 0.973&[0.970, 0.976]\\
				3 &\cmark&\cmark&&& 0.091 & 0.323 & 0.039 & 0.928\\	
				4 &\cmark&&\cmark&& 0.091 & 0.326 & 0.039 & 0.926\\
				5 &\cmark&\cmark&\cmark&& 0.089& 0.319 & \textbf{0.038} & 0.931 \\
				6 &&&&\cmark& 0.091& 0.322 & 0.039 & \textbf{0.934} \\
				7 &\cmark&\cmark&\cmark&\cmark& \textbf{0.088}& \textbf{0.316} & \textbf{0.038} & 0.933 \\
				\Xhline{1.2pt}			
		\end{tabular}}
	\end{center}
	\caption{\textbf{Ablation study of the proposed URCDC-Depth on the NYU-Depth-v2 dataset}. }
	\label{tb:nyudepth}
\end{table}
\begin{figure*}[hbpt]
	\centering
	\includegraphics[width=0.92\linewidth]{nyu_comparison.pdf}% 1\linewidth
	\caption{\textbf{Qualitative depth results on the NYU-Depth-v2 dataset}. The white boxes indicate the regions to emphasize.}
	\label{Fig6}
\end{figure*}
\fi

\begin{table}[hbpt]
\caption{Quantitative depth comparison on the NYU-Depth-v2 dataset with official test split. Metrics follow previous works. }
\renewcommand{\arraystretch}{1.3}
 \setlength\tabcolsep{2.8pt}
\resizebox{1.0\columnwidth}{!}{
\begin{tabular}{c|ccc|ccc}
\Xhline{1.2pt} \textbf{Method} & \textbf{Rel.} $\downarrow$ & \textbf{RMSE} $\downarrow$ & \textbf{log}${_{\bm{{10}}}}$ $\downarrow$ & $\delta^1$ $\uparrow$ & $\delta ^2$ $\uparrow$ & $\delta ^3$ $\uparrow$ \\
\hline
VNL~\cite{yin2019enforcing} & 0.108 & 0.416 & 0.048 & 0.875 & 0.976 & 0.994 \\
BTS~\cite{lee2019big} & 0.113 & 0.407 & 0.049 & 0.871 & 0.977 & 0.995 \\
DAV~\cite{huynh2020guiding} & 0.108 & 0.412 & - & 0.882 & 0.980 & 0.996 \\
PWA~\cite{lee2021patch} & 0.105 & 0.374 & 0.045 & 0.892 & 0.985 & 0.997 \\
TransDepth~\cite{yang2021transformer} & 0.106 & 0.365 & 0.045 & 0.900 & 0.983 & 0.996 \\
Adabins~\cite{bhat2021adabins} & 0.103 & 0.364 & 0.044 & 0.903 & 0.984 & 0.997 \\
P3Depth~\cite{patil2022p3depth} & 0.104 & 0.356 & 0.043 & 0.898 & 0.981 & 0.996 \\
DepthFormer~\cite{li2022depthformer} & 0.096 & 0.339 & 0.041 & 0.921 & 0.989 & 0.998 \\
NeWCRFs~\cite{patil2022p3depth} & 0.095 & 0.334 & 0.041 & 0.922 & \textbf{0.992} & {0.998} \\
PixelFormer~\cite{agarwal2023attention} & 0.090 & 0.322 & 0.039 & 0.929 & 0.991 &  0.998\\
BinsFormer~\cite{li2022binsformer} & 0.094 & 0.330 & 0.040 & 0.925 & 0.989 &  0.997\\
{URCDC-Depth}~\cite{shao2023urcdc} & {0.088} & {0.316} & {0.038} & {0.933} & \textbf{0.992} & {0.998}  \\
VA-Depth~\cite{liu2023va} &  \textbf{0.086} & 0.304 & - & 0.937 & \textbf{0.992} & - \\ \hline
\textbf{DiffusiongDepth} & \textbf{0.085} & \textbf{0.295} & \textbf{0.036} & \textbf{0.939} & \textbf{0.992} & \textbf{0.999} \\
\Xhline{1.2pt}
\end{tabular}
 }
\end{table}

%% file: tables/diffusionproperty.tex
\begin{table}[t]
\caption{\textbf{Ablation study on different inference settings} on NYU-Depth-V2 dataset, t denotes the inference step.\label{tb:disproperties}}
\renewcommand{\arraystretch}{1.3}
 \setlength\tabcolsep{2.8pt}
\resizebox{1.0\columnwidth}{!}{
\begin{tabular}{c|ccc|ccc}
\Xhline{1.2pt} \textbf{Method} & \textbf{Rel.} $\downarrow$ & \textbf{RMSE} $\downarrow$ & \textbf{MAE} $\downarrow$ & $\delta^1$ $\uparrow$ & $\delta ^2$ $\uparrow$ & $\delta ^3$ $\uparrow$ \\
\hline
\textbf{t=20} & \textbf{0.086} & \textbf{0.298} & \textbf{0.166} & \textbf{0.937} & \textbf{0.992} & \textbf{0.999} \\
\hline
\multicolumn{7}{c}{Directly change inference without training.}\\
\hline
t=15 & {0.1178} & {0.4552} & {0.3294} & {0.8644} & {0.9730} & {0.9928} \\
t=10 & { 0.1821} & {0.7506} & {0.5893} & {0.6475} & {0.9289} & {0.9853} \\
t=5 & {0.2873} & {1.1750} & {0.9451} & {0.3803} & {0.7085} & {0.8825} \\
t=2 & {0.3620} & {1.4328} & {1.1616} & {0.2808} & {0.5504} & {0.7699} \\

% t=20 & {0.086} & {0.298} & {0.036} & {0.937} & {0.992} & {0.999} \\
\hline
\multicolumn{7}{c}{Train with different inference steps.}\\
\hline
t=15 & {0.1034} & {0.3648} & {0.238} & {0.9022} & {0.9834} & {0.993} \\
t=10 & {0.1069} & {0.3708} & {0.278} & {0.8815} & {0.9812} & {0.993} \\
t=5 & {0.1108} & {0.4366} & {0.294} & {0.8345} & {0.9644} & {0.992} \\
t=2 & {0.1308} & {0.5678} & {0.387} & {0.8016} & {0.9516} & {0.990} \\
\Xhline{1.2pt}
\end{tabular}
 }
\end{table}

%% file: tables/diffusionprocess.tex
\begin{table}[hbpt]
\caption{\textbf{Ablation on different diffusion methods.}\label{tb:diffusionmethod} Both methods are evaluated on offline official splits.}
\renewcommand{\arraystretch}{1.3}
 \setlength\tabcolsep{2.8pt}
\resizebox{1.0\columnwidth}{!}{
\begin{tabular}{c|ccc|ccc}
\Xhline{1.2pt} \textbf{Method} & \textbf{Rel.} $\downarrow$ & \textbf{RMSE} $\downarrow$ & \textbf{MAE} $\downarrow$ & $\delta^1$ $\uparrow$ & $\delta ^2$ $\uparrow$ & $\delta ^3$ $\uparrow$ \\
\hline
\multicolumn{7}{c}{KITTI Dataset}\\
\hline
\textbf{Refined} & {0.0410} & {1.4523} & {0.7364} & {0.986 } & {0.999} & {1.000} \\
GT & { 0.3480} & { 12.3772} & {7.0154} & {0.4920} & {0.6894} & { 0.8074} \\

% t=20 & {0.086} & {0.298} & {0.036} & {0.937} & {0.992} & {0.999} \\
\hline
\multicolumn{7}{c}{NYU-Depth-V2 Dataset}\\
\hline
\textbf{Refined} & {0.0862} & {0.2983} & {0.1665} & {0.937} & {0.992} & {0.999} \\
GT & { 0.0940} & {0.3041} & {0.1742} & {0.932} & {0.992} & {0.999} \\
\Xhline{1.2pt}
\end{tabular}
 }
\end{table}

%% file: tables/ablationlatent.tex
\begin{table}[hbpt]
\caption{\textbf{Ablation on different depth encoder-decoders and visual conditions} on KITTI Dataset, official offline split (0-50m), where \textbf{DSR} denotes the down-sampling rate of the encoded depth latent. \label{tb:ablationstructure}}
\renewcommand{\arraystretch}{1.3}
 \setlength\tabcolsep{2.8pt}
\resizebox{1.0\columnwidth}{!}{
\begin{tabular}{c|ccc|ccc}
\Xhline{1.2pt} \textbf{Condition} & \textbf{DSR} & \textbf{Rel.} $\downarrow$ & \textbf{RMSE} $\downarrow$ &  $\delta^1$ $\uparrow$ & $\delta ^2$ $\uparrow$ & $\delta ^3$ $\uparrow$ \\
\hline
\multicolumn{7}{c}{Depth Latent Space (Down-Sampling Rate)}\\
\hline
Swin+HAHI & $\bm{\times} \bm{2}$ & {0.0410} & {1.4523} &  {0.986 } & {0.999} & {1.000} \\
Swin+HAHI & $\times 4$ & { 0.0445} & { 1.508} &  {0.985} & {0.999} & { 1.000} \\

% t=20 & {0.086} & {0.298} & {0.036} & {0.937} & {0.992} & {0.999} \\
\Xhline{0.8pt}
\multicolumn{7}{c}{Visual Conditions (Backbones)}\\
\hline
Res34+FPN & $\times 2$ & {0.0554} & {1.7902} & {0.978} & {0.992} & {0.999} \\
Res50+FPN & $\times 2$ & {0.0532} & {1.7124} & {0.978} & {0.993} & {0.999} \\
Swin+FPN & $\times 2$& {0.0458} & { 1.5569} & {0.985} & {0.998} & {0.999} \\
Swin+Bins & $\times 2$ & {0.0468} & {1.5832} & {0.985} & {0.998} & {0.999} \\
\textbf{Swin+HAHI} & $\times 2$ & {0.0410} & {1.4523} & {0.986 } & {0.999} & {1.000} \\
\Xhline{1.2pt}
\end{tabular}
 }
\end{table}

%% file: text/conclusion.tex
\section{Conclusion}

In this paper, we reformulate the monocular depth estimation problem as a diffusion-denoising approach. The iterative refinement of the depth latent helps DiffusionDepth generate accurate and highly detailed depth maps. Experimental results suggest the proposed model reaches state-of-the-art performance. 
This paper verifies the feasibility of introducing a diffusion-denoising model into 3D perception tasks.
A detailed ablation study is provided to give insights for follow-up works. 

%% file: text/appendix.tex
\twocolumn[{%
 \centering
 \section*{\Large \centering Supplementary Material for Submission 4041\\ 
\emph{DiffusionDepth: Diffusion Denoising Approach for Monocular Depth Estimation}}
 % \LARGE The Title\\[1.5em]
 % \large Author: Anton van der Vegt\\[1em]
 \vspace{20pt}
}]
\section{Formulation of Diffusion Model}\label{appendix:diffusion_formula}

%\textbf{Diffusion Process}: 
We define $\bm{x}_{0}$ as the desired refined depth latent, and $\bm{x}_t$ as the depth latent distribution by adding Gaussian noise distribution sequentially $t$ times. 
Following the notion of~\cite{ho2020denoising, dhariwal2021diffusion, nichol2021improved}, this is called the diffusion process. 
Thus, we could continuously add noise into original $\bm{x}_{0}$ through a Markov process sampling variables $\{\bm{x}_0,\bm{x}_1,...\bm{x}_{t-1},\bm{x}_{t},...,\bm{x}_T\}$ until $\bm{x}_T$ becomes a normal noise distribution $p(\bm{x}_T)\sim \mathcal{N}(\bm{x}_T;0,I)$. 
Here, we call the $\bm{x}_T$ as the initialization of the 
% Here, the transition is also called \textit{diffusion process or forward process} as below.
Starting from a refined depth latent distribution $\bm{x}_0$, we define a forward Markovian noising process $q$ as below. In particular, the added noise is scheduled by the variance $\beta_t \bm(I)n (0, 1)$:
\begin{alignat}{2}
    q(\bm{x}_{1:T}|\bm{x}_0) &\coloneqq \prod^T_{t=1}q(\bm{x}_t | \bm{x}_{t-1}) \\
    q(\bm{x}_t | \bm{x}_{t-1}) &\coloneqq \mathcal{N}(\bm{x}_t; \sqrt{1-\beta_t}\bm{x}_{t-1}, \beta_t \bm{I})
\end{alignat}

\noindent As noted by Ho~\etal~\cite{ho2020denoising} of the sampling properties, we can directly sample data $\bm{x}_t$ at an arbitrary timestep $t$ without the need of applying $q$ repeatedly:

\begin{alignat}{2}
    q(\bm{x}_t|\bm{x}_0) &\coloneqq \mathcal{N}(\bm{x}_t; \sqrt{\bar{\alpha}_t}\bm{x}_0, (1 - \bar{\alpha}_t)\bm{I}) \\
    &\coloneqq \sqrt{\bar{\alpha}_t} \bm{x}_0  + \epsilon \sqrt{1 - \bar{\alpha}_t}, \epsilon \bm(I) \mathcal{N}(0, \bm{I})\label{eq:apdiffuse}
\end{alignat}
\noindent where $\bar{\alpha}_t \coloneqq \prod_{s=0}^{t} \alpha_s$ and $\alpha_t \coloneqq 1 - \beta_t$ are also a fixed variance coefficient schedule corresponding to $\beta_t$.  
Based on Bayes' theorem, it is found that the posterior $q(\bm{x}_{t-1}|\bm{x}_t, \bm{x}_0)$ is a Gaussian distribution as well:

\begin{alignat}{2}
     q(\bm{x}_{t-1}|\bm{x}_t, \bm{x}_0) &= \mathcal{N}(\bm{x}_{t-1}; \tilde{\mu}(\bm{x}_t, \bm{x}_0), \tilde{\beta}_t \mathbf{I})\label{eq:posterior}
\end{alignat}

\noindent where
\begin{alignat}{2}
    \tilde{\mu}_t(\bm{x}_t, \bm{x}_0) &\coloneqq
    \frac{\sqrt{\bar{\alpha}_{t-1}}\beta_t}{1-\bar{\alpha}_t}\bm{x}_0 + \frac{\sqrt{\alpha_t}(1-\bar{\alpha}_{t-1})}{1-\bar{\alpha}_t} \bm{x}_t \label{eq:mutilde}
\end{alignat}
\noindent and
\begin{alignat}{2}
    \tilde{\beta}_t &\coloneqq \frac{1-\bar{\alpha}_{t-1}}{1-\bar{\alpha}_t} \beta_t \label{eq:betatilde}
\end{alignat}
 
\noindent are the mean and variance of this Gaussian distribution.

In practice, the representation of $\bm{x}_t$ could be obtained by extending the diffusion process defined in Equation~\ref{eq:apdiffuse} as below.
\begin{equation}\label{eq:diffusionx}
\bm{x}_t = \bar{\alpha}_t\bm{x}_0 + \alpha_t\bar{\beta}_{t-1}\bar{\boldsymbol{\vartheta}}_{t-1} + \beta_t \boldsymbol{\vartheta}_t \end{equation}
where $\boldsymbol{\vartheta_t} \sim \mathcal{N}(0,\bm(I))$ is a gaussian distribution that represents the stochastic property of the diffusion process. It also gives a description of how to represent the diffusion result in $\bm{x}_t$ by real sample $\bm{x}_0$ and given fixed variance scheduler $\alpha_t$ and $\beta_t$.
We could get a sample from $q(\bm{x}_0)$ by first sampling from $q(\bm{x}_T)$ and running the reversing steps $q(\bm{x}_{t-1} | \bm{x}_t)$ until $\bm{x}_0$. 

Besides, the distribution of $q(\bm{x}_T)$ is nearly an isotropic Gaussian distribution with a sufficiently large $T$ and reasonable schedule of $\beta_t$~($\beta_t \rightarrow 0$), which making it trivial to sample $\bm{x}_T \sim \mathcal{N}(0, \bm{I})$. Moreover, since calculating $q(\bm{x}_{t-1} | \bm{x}_t)$ exactly should depend on the entire data distribution, we could approximate $q_{\theta}(\bm{x}_{t-1} | \bm{x}_t)$ using a neural network posterior process,
which is optimized to predict a mean $\mu_\theta$ and a diagonal covariance matrix $\Sigma_\theta$:
\begin{alignat}{2}
p_{\theta}(\bm{x}_{t-1}|\bm{x}_t) &\coloneqq \mathcal{N}(\bm{x}_{t-1};\mu_{\theta}(\bm{x}_t, t), \sigma_{\theta}(\bm{x}_t, t)) \label{eq:ptheta_ap}
\end{alignat}
Instead of directly parameterizing $\mu_\theta(\bm{x}_t, t)$, Ho~\etal~\cite{ho2020denoising} found learning a network $f_\theta(\bm{x}_t, t)$ to predict the $\epsilon$ or $\bm{x}_0$ from \ref{eq:apdiffuse} worked best. We choose to predict $\bm{x}_0$ in this work.

However, depth denoising can't be performed without any given images. We utilize the visual condition $c$ to guide the denoising process. 
As we assume above, the condition, $\bm{c}$ is entirely independent of the denoising and diffusion process. 
So we can easily change it into conditioned denoising by simply modifying the denoising process as below. 
\begin{alignat}{2}
p_{\theta}(\bm{x}_{t-1}|\bm{x}_t,\bm{c}) &\coloneqq \mathcal{N}(\bm{x}_{t-1};\mu_{\theta}(\bm{x}_t, t, \bm{c}), \sigma_{\theta}(\bm{x}_t, t)) \label{eq:ptheta_ap2}
\end{alignat}

However, since we desired a deterministic process, which means that given an image, the generated depth map is uniquely deterministic. This is a pre-request for accurate prediction tasks.
In that case, we eliminate randomness according to DDIM~\cite{song2020improved}. Where the random variance $\boldsymbol{\vartheta_t}$ is set to 0.

 \section{Extended Implementation details}
The implementation and the pre-trained model will be open-source after acceptance. 
% Here we release an anonymous link~\footnote{\href{https://anonymous.4open.science/r/DiffusionDepth-4951}{https://anonymous.4open.science/r/DiffusionDepth-4951}} for review.  
DiffusionDepth is implemented with the Pytorch~\cite{paszke2019pytorch} framework. We train the entire model with batch size 16 for 30 epochs iterations on a single node with 8 NVIDIA A100 40G GPUs. 
We utilize the AdamW optimizer~\cite{kingma2014adam} with ($\beta_1$, $\beta_2$, $w$) = (0.9, 0.999, 0.01), where $w$ is the weight decay. 
The linear learning rate warm-up strategy is applied for the first $15\%$ iterations. The cosine annealing learning rate strategy is adopted for the learning rate decay from the initial learning rate of $1e-4$ to $1e-8$. 
We use L1 and L2 pixel-wise depth loss at the first $50\%$ training iterations as auxiliary subversion. 
% For the NYU-Depth-v2 dataset, we utilize the official 25 classes divided by folder names for the auxiliary scene understanding task. For KITTI, since the outdoor dataset is tough to classify, we omit the scene classification loss and only use ground truth depth to provide supervision.
\paragraph{Augmentation}
We randomly crop a patch from the original image and its corresponding depth map and resize them to the desired input size. This helps to avoid overfitting and focus on learning to refine the details of different regions of the image.
For the KITTI dataset, we sequentially utilize the random crop with size $706\times352$, color jitter with various lightness saturation, random scale from $1.0$ to $1.5$ times, and random flip for training data augmentation. 
For the NYU-Depth-V2 dataset, we use the same augmentation with the random crop with size $512\times340$. 
We randomly adjust the brightness, contrast, saturation, and hue of the input image. This helps to simulate different lighting conditions and make the network invariant with color changes. 
We randomly flip the input image and its corresponding depth map horizontally. This helps to augment the data with different orientations and symmetries.
Random rotation: We randomly rotate the input image and its corresponding depth map by a small angle. This helps to augment the data with different angles and perspectives. Here, we use $-5-5$ degrees as the rotation parameter.

\paragraph{Visual Condition:} DiffusionDepth is compatible with any backbone which could extract multi-scale features. 
Here, we respectively evaluate our model on the standard convolution-based ResNet~\cite{he2016deep} backbones and transformer-based Swin~\cite{liu2021swin} backbones. 
We employ hierarchical aggregation and heterogeneous interaction (HAHI~\cite{li2022depthformer}) neck to enhance features between scales and feature pyramid neck~\cite{lin2017feature} to aggregate features into monocular visual condition. 
The visual condition dimension is equal to the last layer of the neck. 
We respectively use channel dimensions $[64,128,256,512]$ and $[192, 384, 768, 1536]$
for ResNet and Transformer backbones. 

\paragraph{Diffusiong Head:} We use the improved sampling process~\cite{song2020improved} with 1000 diffusion steps for training and 20 inference steps for inference. 
The learning rate of the diffusion head is 10 times larger than the backbone parameters. 
The dimension $d$ of the encoded depth latent is 16 with shape $\frac{H}{2},\frac{W}{2}, d$, we conduct detailed ablation to illustrate different inference settings. The max depth value of the decoder is $1e6$ for all experiments.

\subsection{Evaluation Metrics}
\label{sec:evaluation_metrics}
Suppose the predicted and ground-truth depth to be $\hat{X}\in\mathbb{R}^{m \times n}$ and ${X}_{gt} \in \mathbb{R}^{m \times n}$, respectively, and the number of valid pixels to be $N$. 
We follow the existing methods~\cite{liu2023va} and utilize the following measures for quantitative evaluation. We listed all potential metrics below: 
\begin{itemize}
\item square root of the Scale Invariant Logarithmic error (\textbf{SILog}):
\begin{equation}
    \frac{1}{N}\sum_{i,j}(e^{i,j})^2 - \frac{1}{N^2}(\sum_{i,j}e^{i,j})^2
\end{equation}
where $e^{i,j} = \log{\hat{X}^{i,j}}-\log{Z_{gt}^{i,j}}$; 
\item Relative Squared error (\textbf{Sq Rel}): 
\begin{equation}
    \frac{1}{N}\sum_{i,j}(\hat{X}^{i,j}-Z_{gt}^{i,j})^2 / Z_{gt}^{i,j}
\end{equation}
\item Relative Absolute Error (\textbf{Abs Rel}): 
\begin{equation}
    \frac{1}{N}\sum_{i,j}|\hat{X}^{i,j}-Z_{gt}^{i,j}| / Z_{gt}^{i,j}
\end{equation}
\item Root Mean Squared error (\textbf{RMS}): 
\begin{equation}
    \frac{1}{N}\sqrt{\sum_{i,j}(\hat{X}^{i,j}-Z_{gt}^{i,j})^2}
\end{equation}
\item Root Mean Squared Logarithmic error (\textbf{RMS log}): 
\begin{equation}
    \frac{1}{N}\sqrt{\sum_{i,j}(e^{i,j})^2}
\end{equation}
\item threshold accuracy ($\bm{\delta_k}$): percentage of $\hat{X}^{i,j}$ s.t. $\max (\frac{\hat{X}^{i,j}}{Z_{gt}^{i,j}}, \frac{Z_{gt}^{i,j}}{\hat{X}^{i,j}}) < 1.25^k$.
\end{itemize}

\section{Extensive Qualitative Comparison on NYU-Depth-V2 Dataset.}

As mentioned in the main paper, the proposed DiffusionDepth reaches state-of-the-art performance on the NYU-Depth-V2 dataset. Since URCDC-depth and VA-Depth have not been open-sourced yet, we select BinsFormer and DepthFormer as the comparison candidates. 
The extended visual comparison is reported in Fig.~\ref{fig:nyudepthnew}.

\begin{figure}[t]
\caption{ Qualitative Comparison on NYU-Depth-V2 Dataset. \label{fig:nyudepthnew}}
\end{figure}

\begin{figure*}[t]
  \centering
 \subfigure[]{
\includegraphics[width=0.47\textwidth]{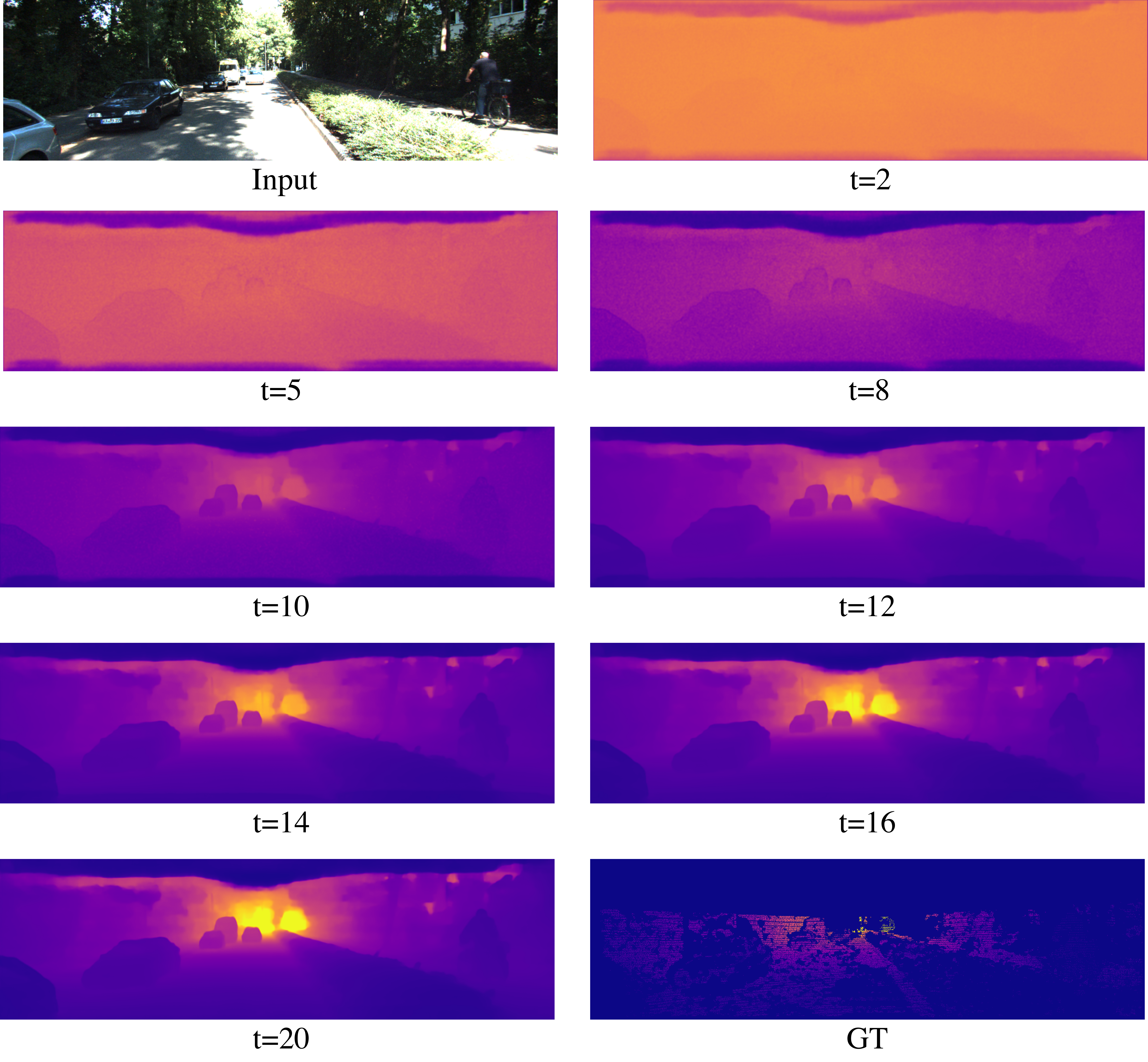}
}
\hfill
 \subfigure[]{
\includegraphics[width=0.47\textwidth]{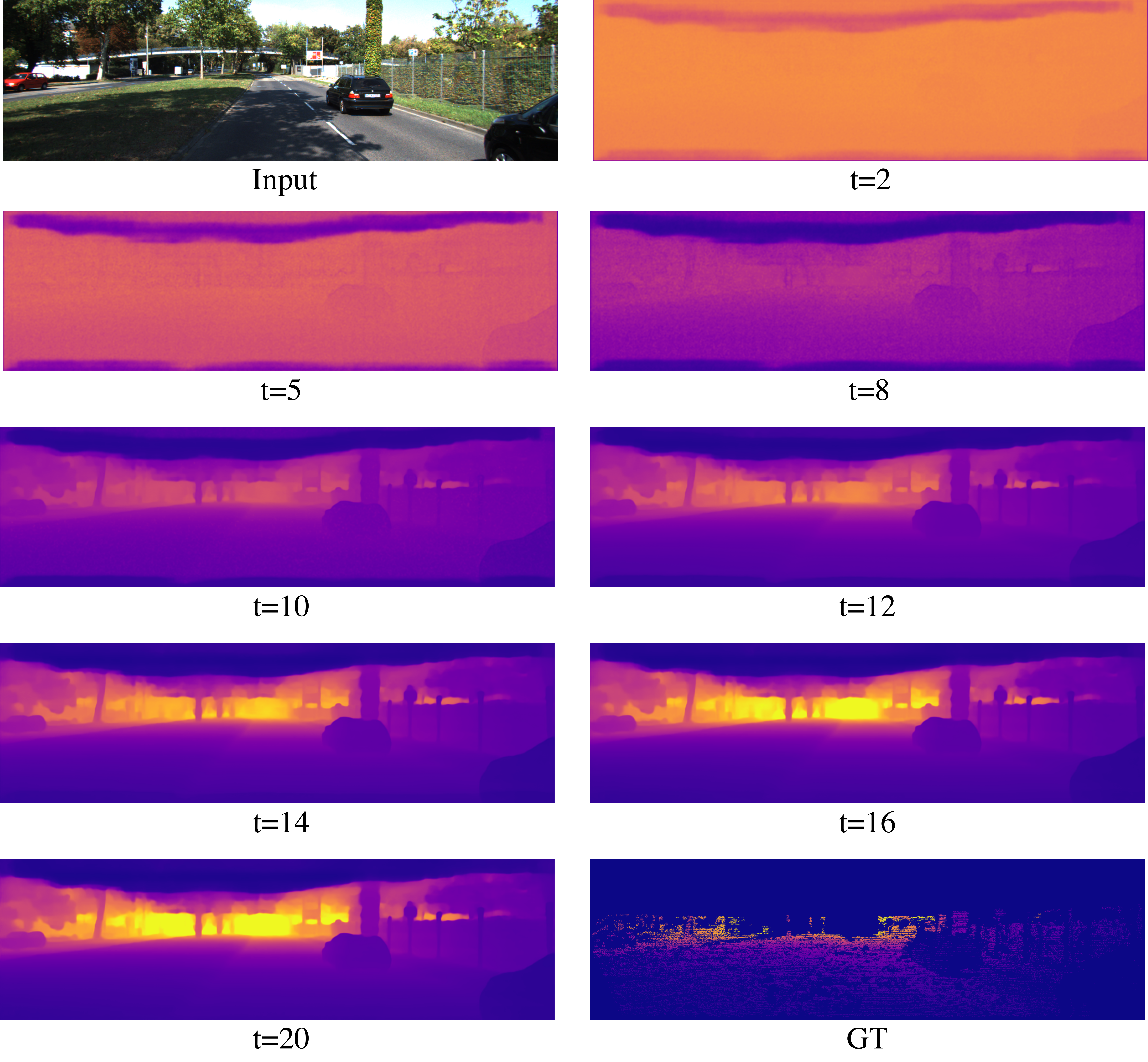}
}
\hfill
 \subfigure[]{
\includegraphics[width=0.47\textwidth]{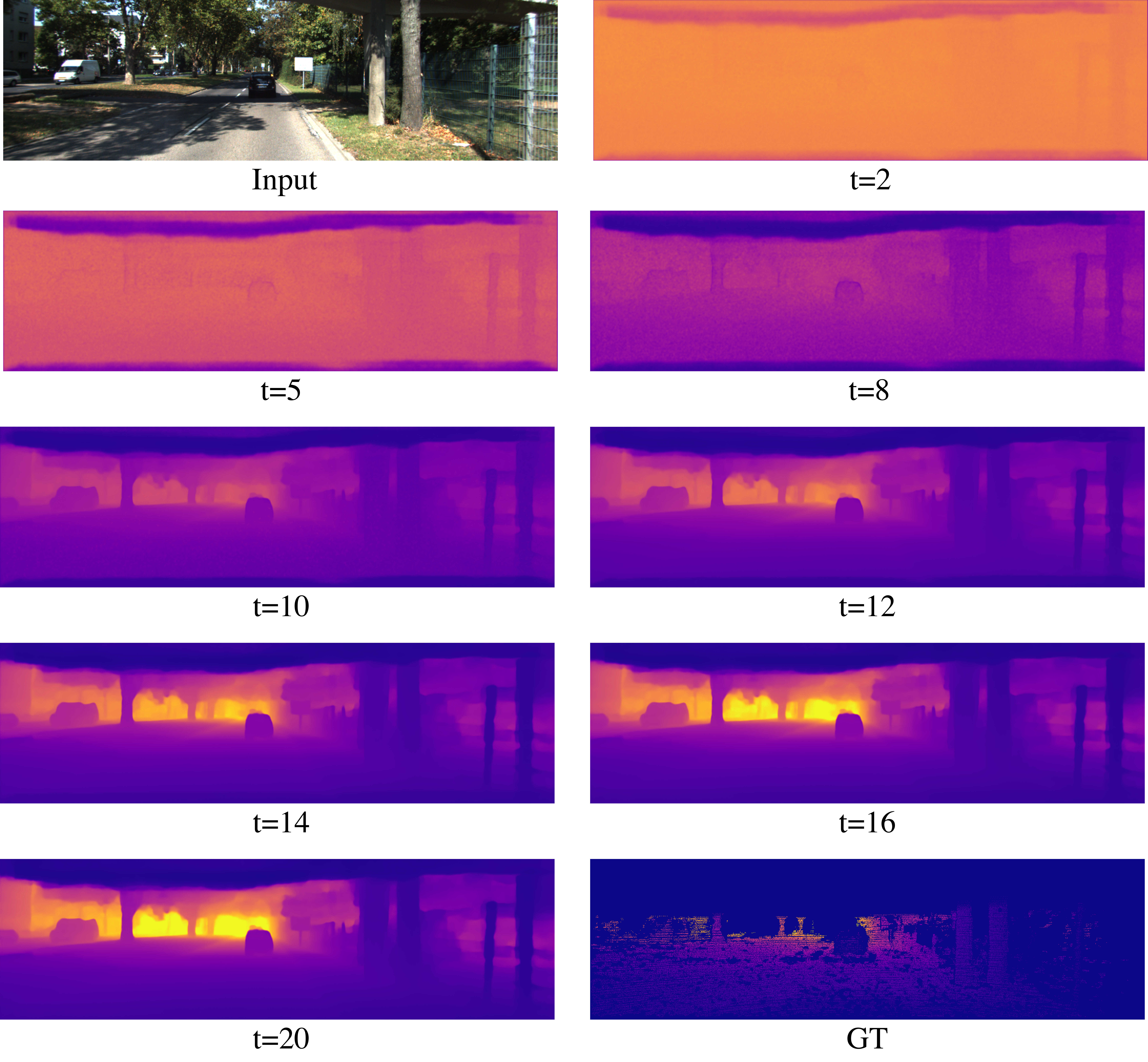}
}
\hfill
 \subfigure[]{
\includegraphics[width=0.47\textwidth]{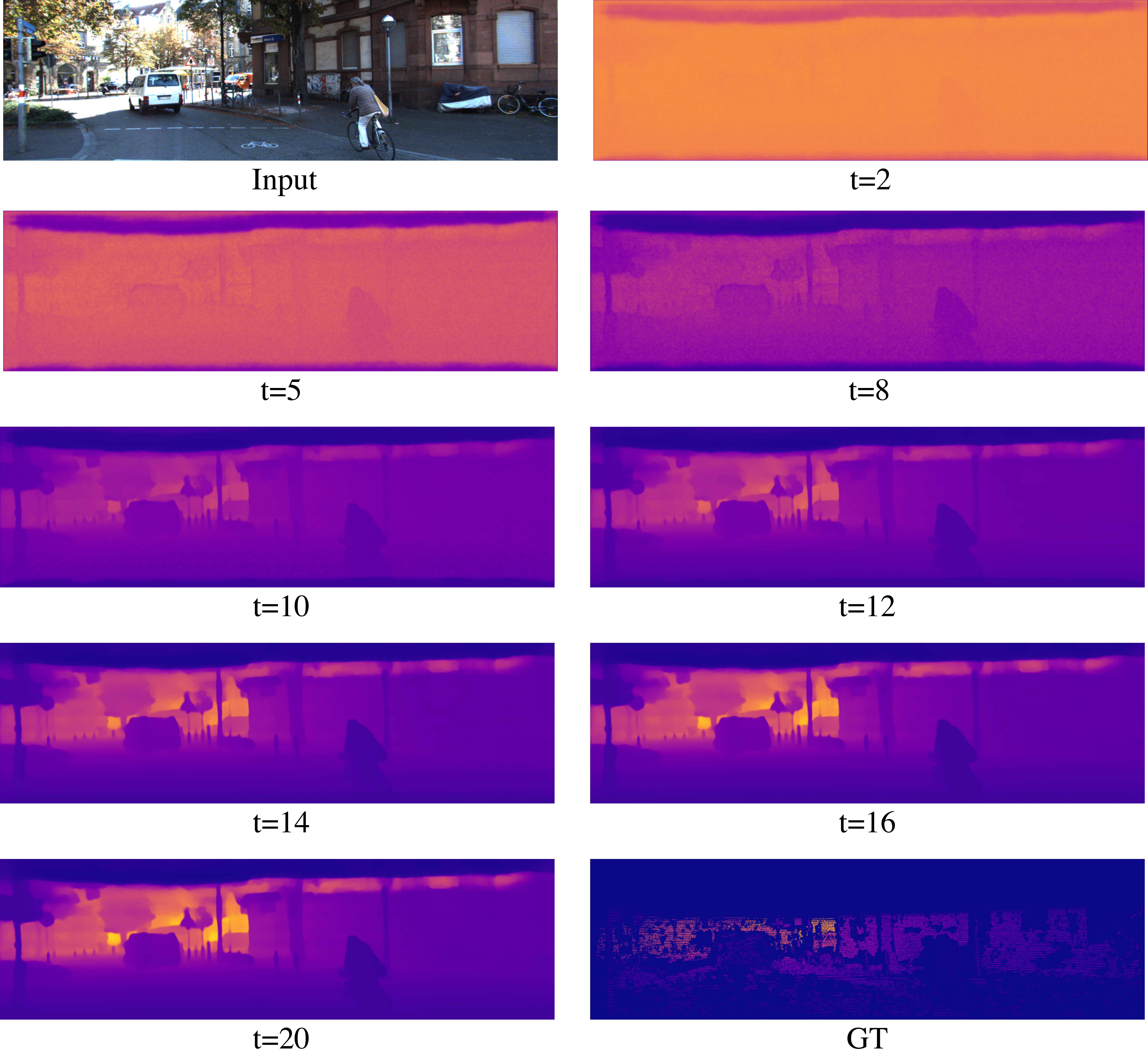}
}
  \caption{Visualization of the diffusion process on KITTI dataset. \label{fig:kitti8}}
\end{figure*}

\begin{figure*}[t]
  \centering
 \subfigure[]{
\includegraphics[width=0.47\textwidth]{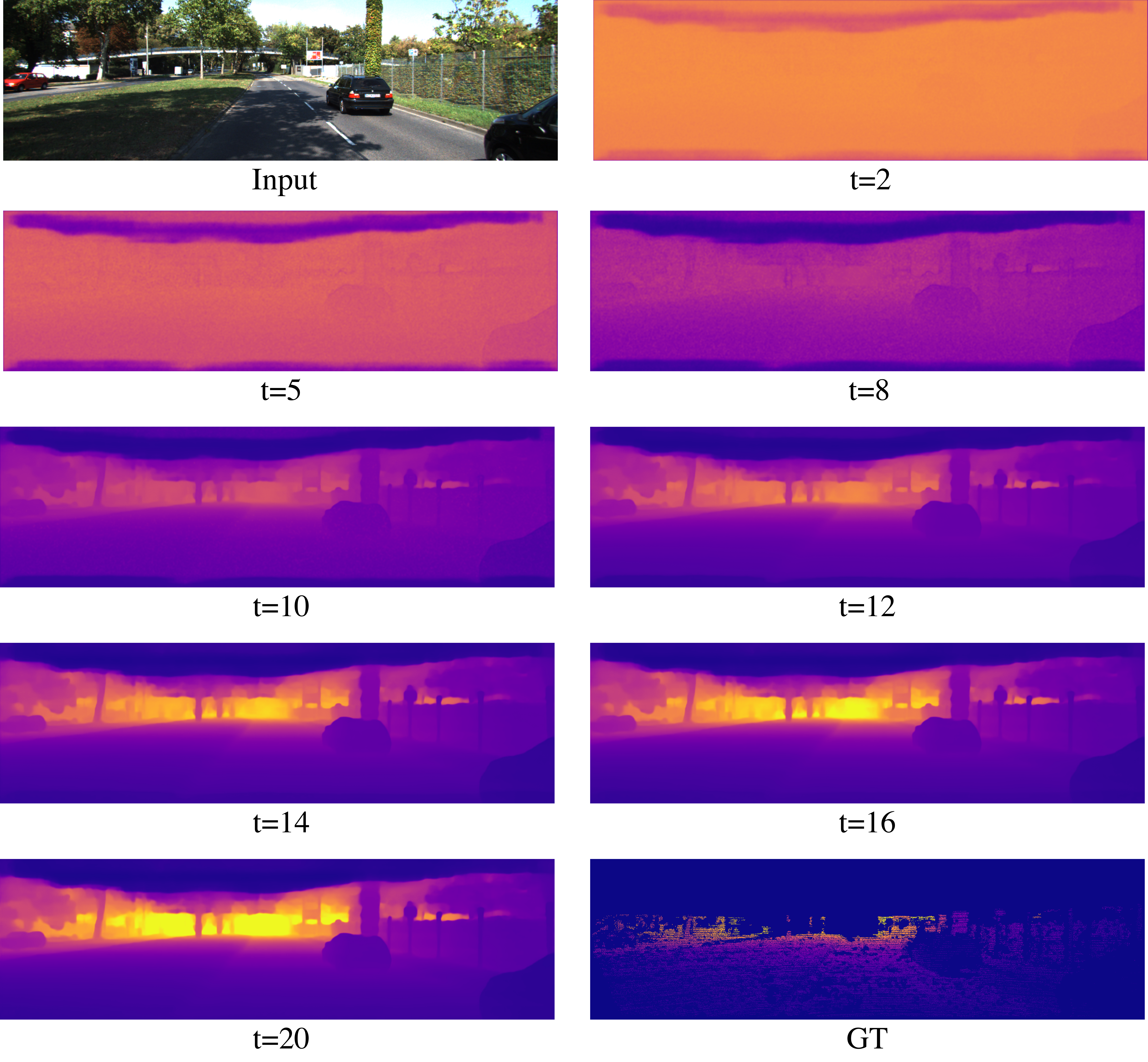}
}
\hfill
 \subfigure[]{
\includegraphics[width=0.47\textwidth]{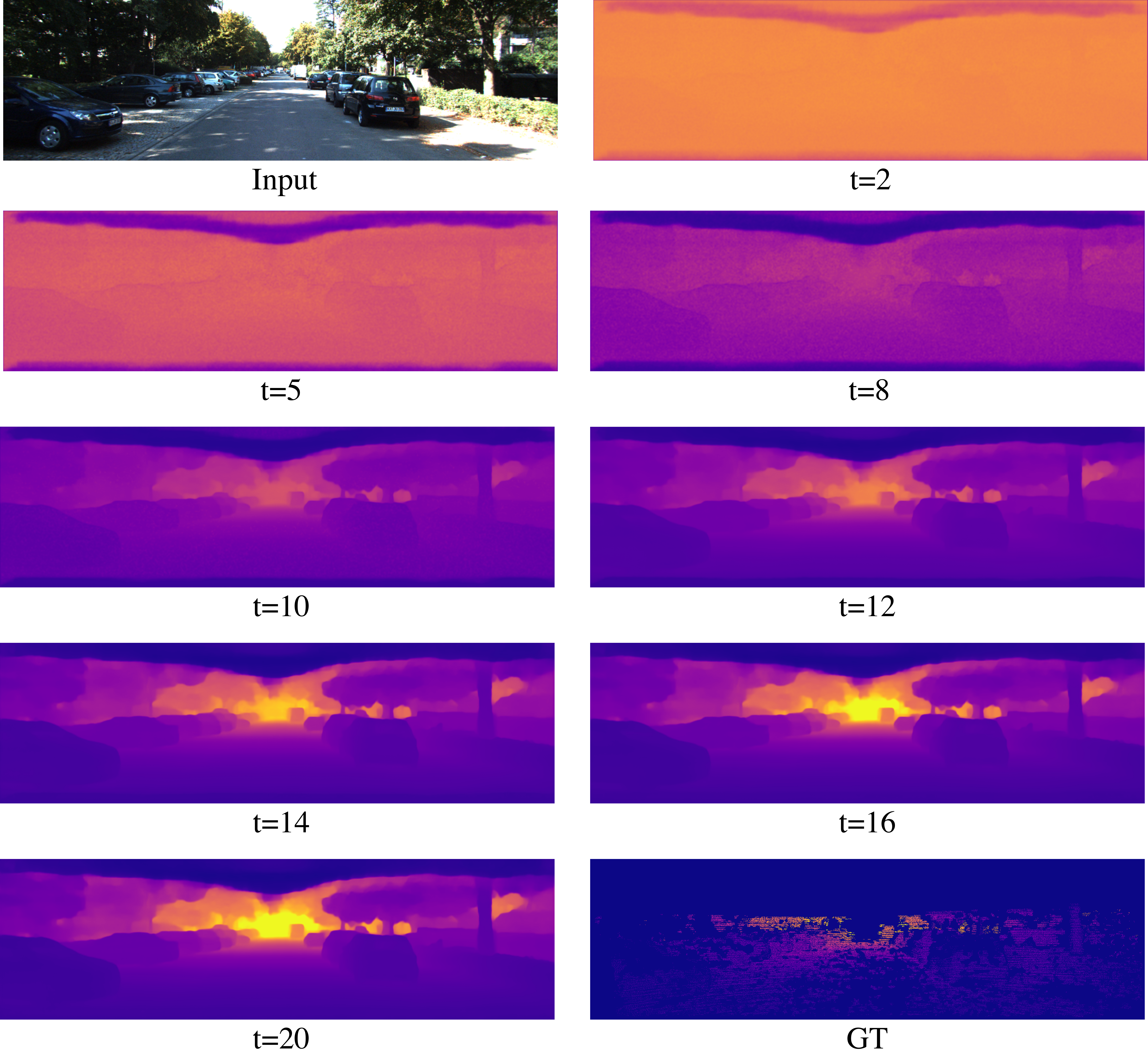}
}
\hfill
 \subfigure[]{
\includegraphics[width=0.47\textwidth]{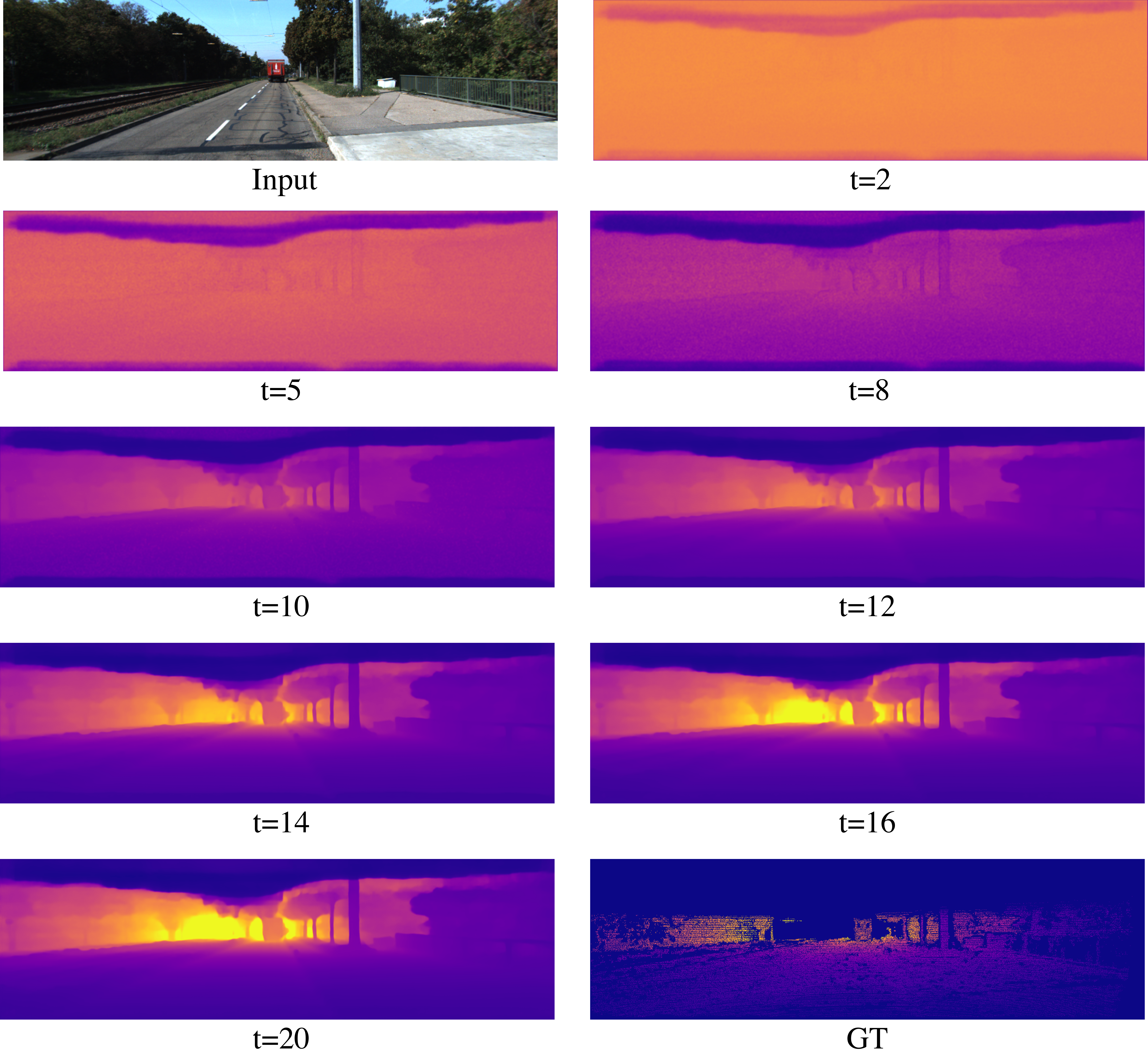}
}
\hfill
 \subfigure[]{
\includegraphics[width=0.47\textwidth]{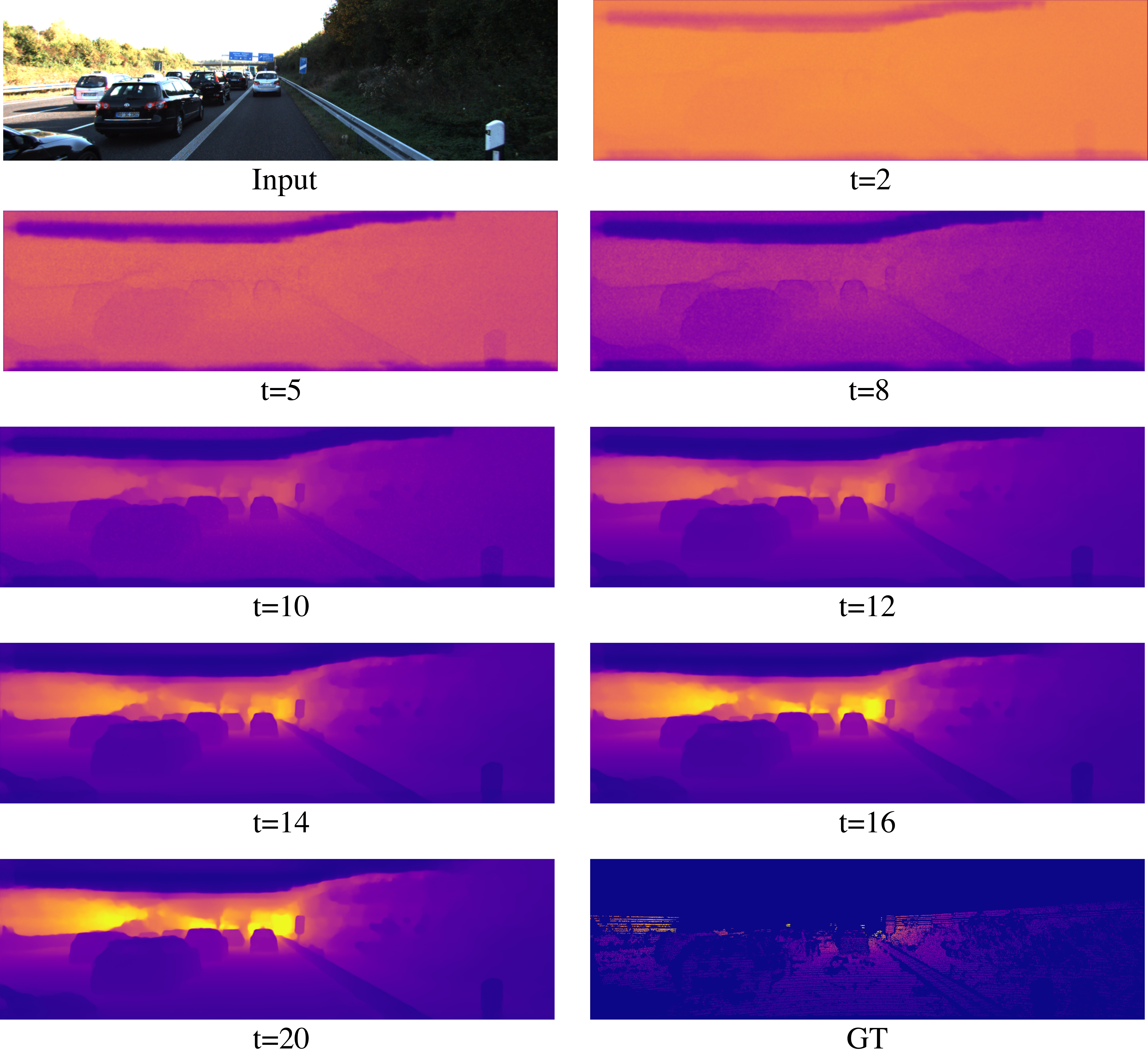}
}
  \caption{Visualization of the diffusion process on KITTI dataset. \label{fig:kitti1}}
\end{figure*}

\section{Illustration of the Denoising Process.}
We present more visualization results on KITTI to illustrate the diffusion process in Fig.~\ref{fig:kitti1} and Fig.~\ref{fig:kitti8}.
The denoising process starts by initializing the upper and lower bound of the predictable area (t=2, t=5). 
We observe a black area at the upper part of the depth map, which corresponds to the end of the road or part of the sky that have very large depth values and are not predicted by our model. 
The subsequent steps (t=8,t=10) refine the shapes and basic \textit{structure} of the whole depth map. 
Then, the denoising model gradually adjusts the depth map to match accurate distance correlations and real depth values. 
Meanwhile, the shape of the predicted objects becomes more sharp and clear through the steps.

\iffalse
\begin{figure}[t]
\includegraphics[width=0.47\textwidth]{img/diffusionprocess/kitti1.pdf}
\caption{Visualization of the diffusion process on KITTI dataset. \label{fig:kitti1}}
\end{figure}
\begin{figure}[t]
\includegraphics[width=0.47\textwidth]{img/diffusionprocess/kitti1.pdf}
\caption{Visualization of the diffusion process on KITTI dataset. \label{fig:kitti2}}
\end{figure}

\begin{figure}[t]
\includegraphics[width=0.47\textwidth]{img/diffusionprocess/kitti3.pdf}
\caption{Visualization of the diffusion process on KITTI dataset.\label{fig:kitti3}}
\end{figure}

\begin{figure}[t]
\includegraphics[width=0.47\textwidth]{img/diffusionprocess/kitti4.pdf}
\caption{Visualization of the diffusion process on KITTI dataset.\label{fig:kitti4}}
\end{figure}
\fi

We present more visualization results on NYU-Depth V2 to illustrate the diffusion process in Fig.~\ref{fig:nyu1}.
As the GT depth is dense in the NYU-Depth-V2 dataset, we do not observe any unpredictable areas as it is in the KITTI dataset. 
The subsequent steps (t<10) refine the shapes and basic \textit{structure} of the whole depth map. 
Then, the denoising model gradually adjusts the depth map to match accurate distance correlations and real depth values. 
Especially, this process could deal with edges' extreme depth variances better than previous baselines.   
Meanwhile, the shape of the predicted objects becomes more sharp and clear through the steps.

\iffalse
\begin{figure}[t]
\includegraphics[width=0.47\textwidth]{img/diffusionprocess/nyu1.pdf}
\caption{Visualization of the diffusion process on NYU-Depth-V2 dataset. \label{fig:nyu1}}
\end{figure}
\begin{figure}[t]
\includegraphics[width=0.47\textwidth]{img/diffusionprocess/nyu2.pdf}
\caption{Visualization of the diffusion process on NYU-Depth-V2 dataset. \label{fig:nyu2}}
\end{figure}

\begin{figure}[t]
\includegraphics[width=0.47\textwidth]{img/diffusionprocess/nyu3.pdf}
\caption{Visualization of the diffusion process on NYU-Depth-V2 dataset.\label{fig:nyu3}}
\end{figure}

\begin{figure}[t]
\includegraphics[width=0.47\textwidth]{img/diffusionprocess/nyu4.pdf}
\caption{Visualization of the diffusion process on NYU-Depth-V2 dataset.\label{fig:nyu4}}
\end{figure}
\fi

\begin{figure*}[t]
  \centering
 \subfigure[]{
\includegraphics[width=0.47\textwidth]{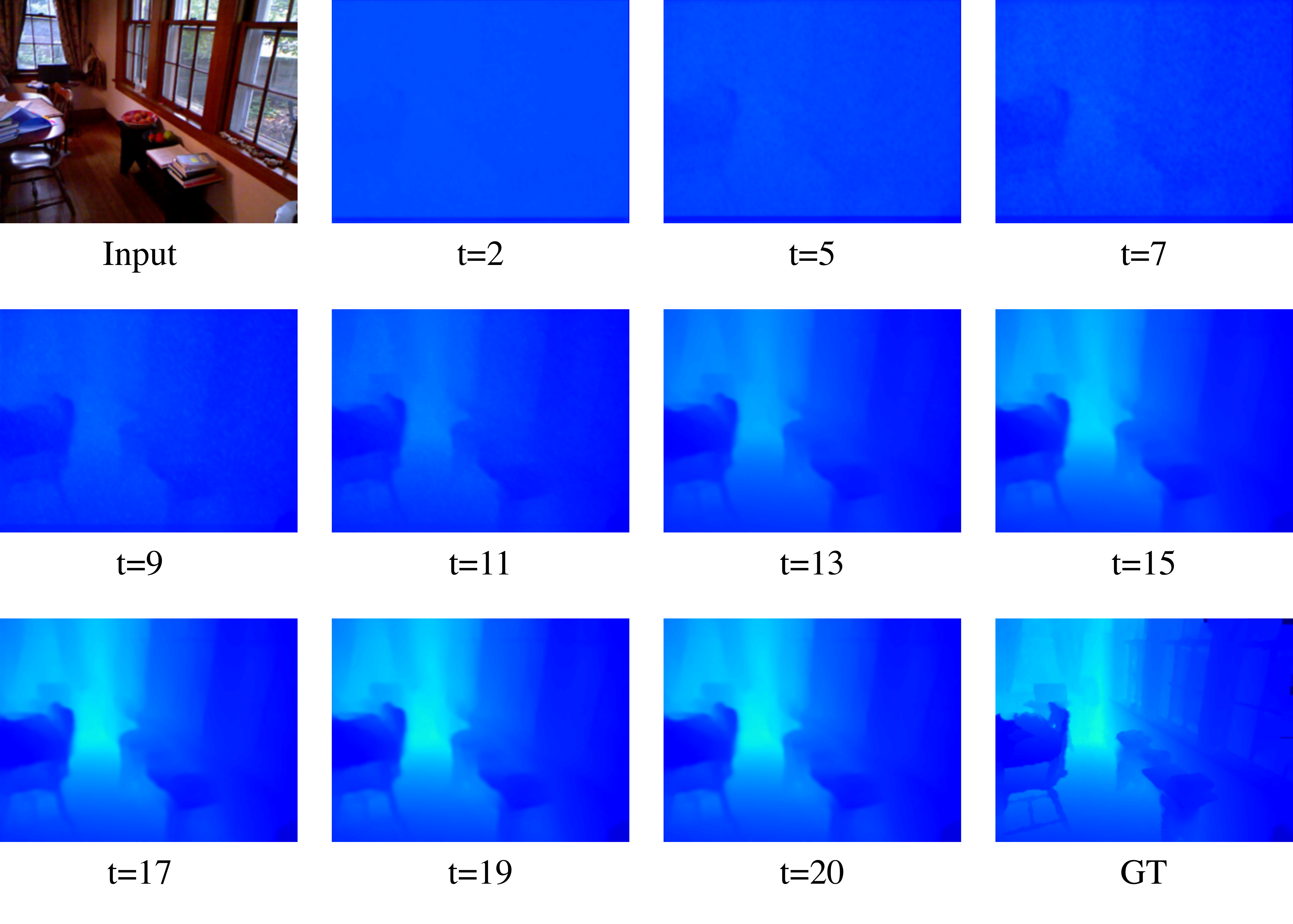}
}
\hfill
 \subfigure[]{
\includegraphics[width=0.47\textwidth]{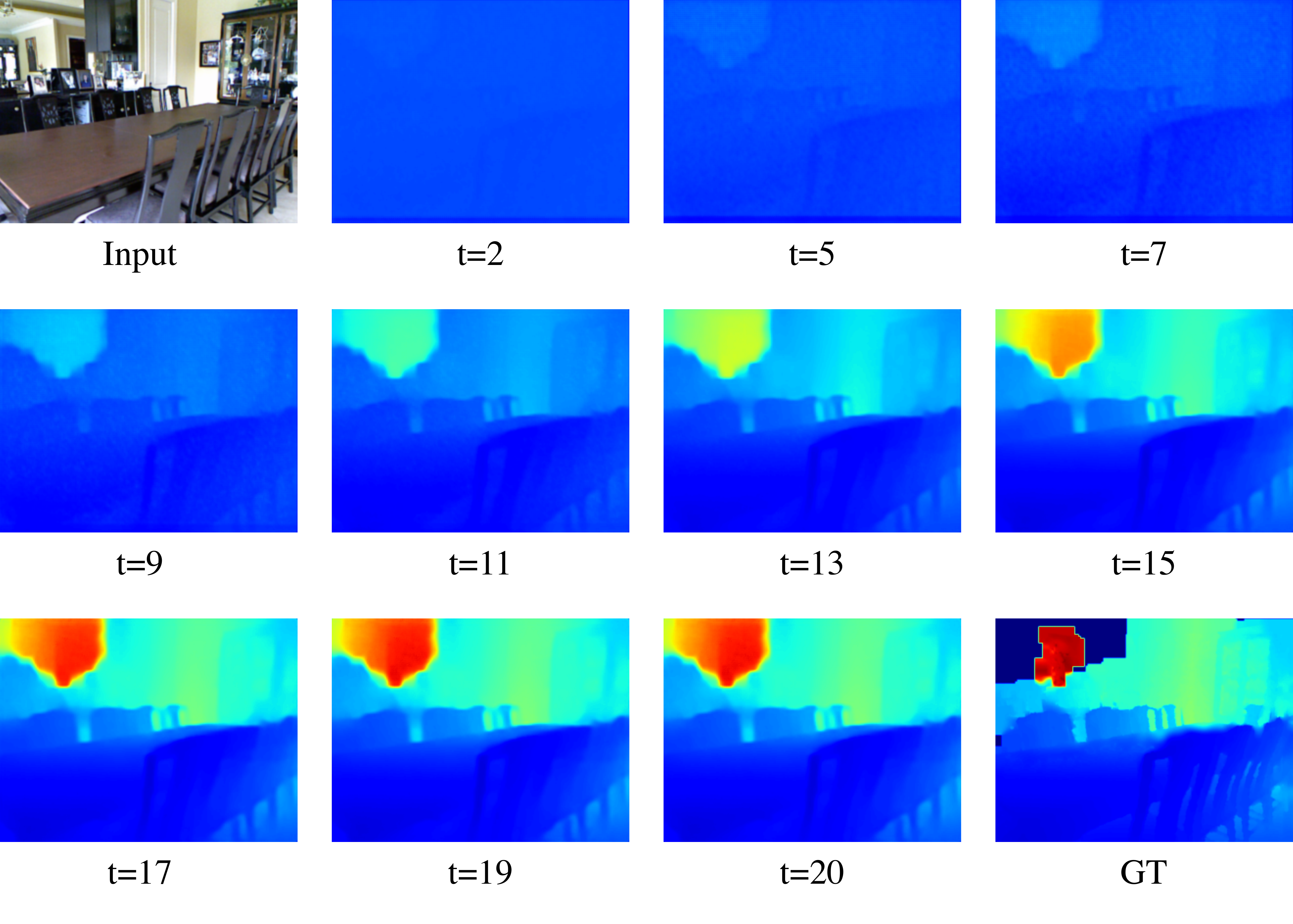}
}
\hfill
 \subfigure[]{
\includegraphics[width=0.47\textwidth]{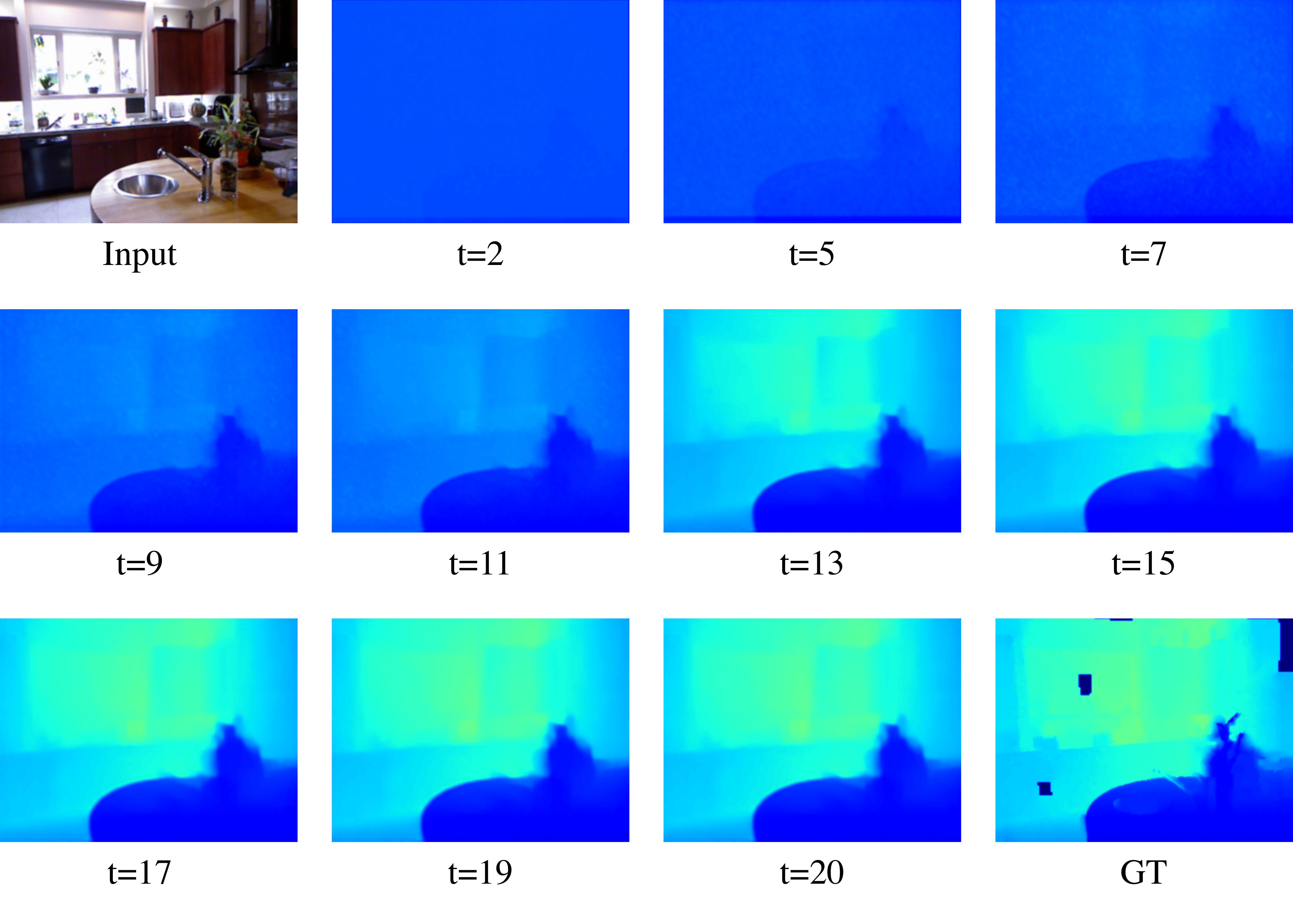}
}
\hfill
 \subfigure[]{
\includegraphics[width=0.47\textwidth]{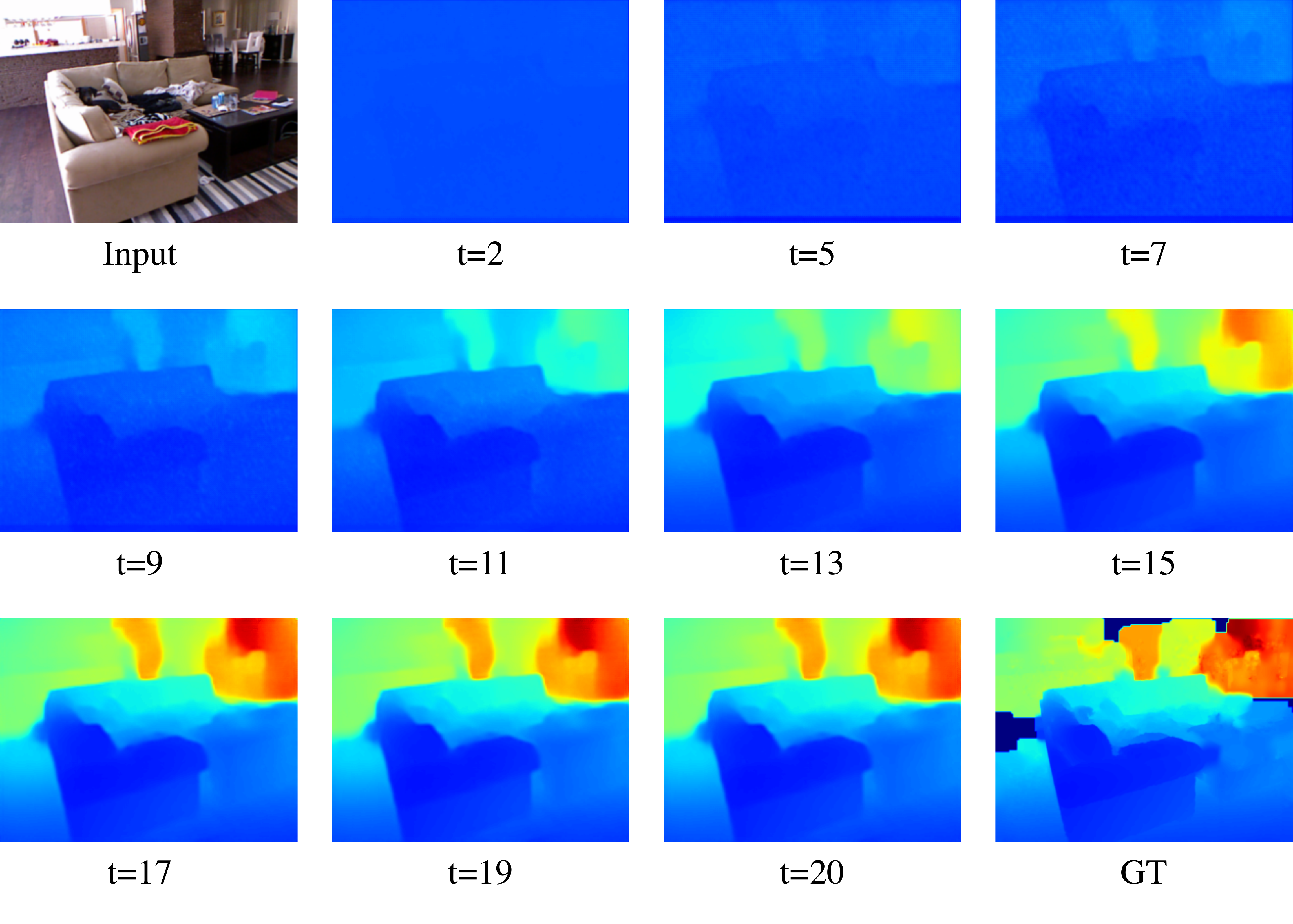}
}
  \caption{Visualization of the diffusion process on NYU-Depth-V2 dataset.\label{fig:nyu1}}
\end{figure*}

% \section{Hyper-Parameters}

%% file: main.bbl
\begin{thebibliography}{10}\itemsep=-1pt

\bibitem{agarwal2023attention}
Ashutosh Agarwal and Chetan Arora.
\newblock Attention attention everywhere: Monocular depth prediction with skip
  attention.
\newblock In {\em Proceedings of the IEEE/CVF Winter Conference on Applications
  of Computer Vision (WACV)}, pages 5861--5870, 2023.

\bibitem{aich2021bidirectional}
Shubhra Aich, Jean Marie~Uwabeza Vianney, Md~Amirul Islam, and Mannat
  Kaur~Bingbing Liu.
\newblock Bidirectional attention network for monocular depth estimation.
\newblock In {\em 2021 IEEE International Conference on Robotics and Automation
  (ICRA)}, pages 11746--11752. IEEE, 2021.

\bibitem{amit2021segdiff}
Tomer Amit, Eliya Nachmani, Tal Shaharbany, and Lior Wolf.
\newblock Segdiff: Image segmentation with diffusion probabilistic models.
\newblock {\em arXiv preprint arXiv:2112.00390}, 2021.

\bibitem{baranchuk2022labelefficient}
Dmitry Baranchuk, Andrey Voynov, Ivan Rubachev, Valentin Khrulkov, and Artem
  Babenko.
\newblock Label-efficient semantic segmentation with diffusion models.
\newblock In {\em International Conference on Learning Representations (ICLR)},
  2022.

\bibitem{bhat2021adabins}
Shariq~Farooq Bhat, Ibraheem Alhashim, and Peter Wonka.
\newblock Adabins: Depth estimation using adaptive bins.
\newblock In {\em Computer Vision and Pattern Recognition (CVPR)}, pages
  4009--4018, 2021.

\bibitem{brempong2022denoising}
Emmanuel~Asiedu Brempong, Simon Kornblith, Ting Chen, Niki Parmar, Matthias
  Minderer, and Mohammad Norouzi.
\newblock Denoising pretraining for semantic segmentation.
\newblock In {\em Proceedings of the IEEE/CVF Conference on Computer Vision and
  Pattern Recognition (CVPR)}, pages 4175--4186, 2022.

\bibitem{chen2022diffusiondet}
Shoufa Chen, Peize Sun, Yibing Song, and Ping Luo.
\newblock Diffusiondet: Diffusion model for object detection.
\newblock {\em arXiv preprint arXiv:2211.09788}, 2022.

\bibitem{chen2022generalist}
Ting Chen, Lala Li, Saurabh Saxena, Geoffrey Hinton, and David~J Fleet.
\newblock A generalist framework for panoptic segmentation of images and
  videos.
\newblock {\em arXiv preprint arXiv:2210.06366}, 2022.

\bibitem{dhariwal2021diffusion}
Prafulla Dhariwal and Alexander Nichol.
\newblock Diffusion models beat gans on image synthesis.
\newblock {\em Advances in neural information processing systems
  (NIPS/NeurIPS)}, 34:8780--8794, 2021.

\bibitem{diaz2019soft}
Raul Diaz and Amit Marathe.
\newblock Soft labels for ordinal regression.
\newblock In {\em Computer Vision and Pattern Recognition (CVPR)}, pages
  4738--4747, 2019.

\bibitem{dosovitskiy2020vit}
Alexey Dosovitskiy, Lucas Beyer, Alexander Kolesnikov, Dirk Weissenborn,
  Xiaohua Zhai, Thomas Unterthiner, Mostafa Dehghani, Matthias Minderer, Georg
  Heigold, Sylvain Gelly, et~al.
\newblock An image is worth 16x16 words: Transformers for image recognition at
  scale.
\newblock {\em arXiv preprint arXiv:2010.11929}, 2020.

\bibitem{duan2019learning}
Yiqun Duan and Chen Feng.
\newblock Learning internal dense but external sparse structures of deep
  convolutional neural network.
\newblock In {\em Artificial Neural Networks and Machine Learning--ICANN 2019:
  Deep Learning: 28th International Conference on Artificial Neural Networks,
  Munich, Germany, September 17--19, 2019, Proceedings, Part II 28}, pages
  247--262. Springer, 2019.

\bibitem{eigen2015predicting}
David Eigen and Rob Fergus.
\newblock Predicting depth, surface normals and semantic labels with a common
  multi-scale convolutional architecture.
\newblock In {\em Proceedings of the IEEE International Conference on Computer
  Vision (ICCV)}, pages 2650--2658, 2015.

\bibitem{eigen2014depth}
David Eigen, Christian Puhrsch, and Rob Fergus.
\newblock Depth map prediction from a single image using a multi-scale deep
  network.
\newblock In {\em Advances in neural information processing systems
  (NIPS/NeurIPS)}, pages 2366--2374, 2014.

\bibitem{fu2018deep}
Huan Fu, Mingming Gong, Chaohui Wang, Kayhan Batmanghelich, and Dacheng Tao.
\newblock Deep ordinal regression network for monocular depth estimation.
\newblock In {\em Proceedings of the IEEE Conference on Computer Vision and
  Pattern Recognition (CVPR)}, pages 2002--2011, 2018.

\bibitem{geiger2013kitti}
Andreas Geiger, Philip Lenz, Christoph Stiller, and Raquel Urtasun.
\newblock Vision meets robotics: The kitti dataset.
\newblock {\em The International Journal of Robotics Research},
  32(11):1231--1237, 2013.

\bibitem{graikos2022diffusion}
Alexandros Graikos, Nikolay Malkin, Nebojsa Jojic, and Dimitris Samaras.
\newblock Diffusion models as plug-and-play priors.
\newblock {\em arXiv preprint arXiv:2206.09012}, 2022.

\bibitem{guizilini2021sparse}
Vitor Guizilini, Rares Ambrus, Wolfram Burgard, and Adrien Gaidon.
\newblock Sparse auxiliary networks for unified monocular depth prediction and
  completion.
\newblock In {\em Proceedings of the IEEE Conference on Computer Vision and
  Pattern Recognition (CVPR)}, pages 11078--11088, 2021.

\bibitem{guo2018learning}
Xiaoyang Guo, Hongsheng Li, Shuai Yi, Jimmy Ren, and Xiaogang Wang.
\newblock Learning monocular depth by distilling cross-domain stereo networks.
\newblock In {\em Proceedings of the European Conference on Computer Vision
  (ECCV)}, pages 484--500, 2018.

\bibitem{He_2016_CVPR}
Kaiming He, Xiangyu Zhang, Shaoqing Ren, and Jian Sun.
\newblock Deep residual learning for image recognition.
\newblock In {\em Proceedings of the IEEE Conference on Computer Vision and
  Pattern Recognition (CVPR)}, June 2016.

\bibitem{he2016deep}
Kaiming He, Xiangyu Zhang, Shaoqing Ren, and Jian Sun.
\newblock Deep residual learning for image recognition.
\newblock In {\em Proceedings of the IEEE Conference on Computer Vision and
  Pattern Recognition (CVPR)}, pages 770--778, 2016.

\bibitem{ho2020denoising}
Jonathan Ho, Ajay Jain, and Pieter Abbeel.
\newblock Denoising diffusion probabilistic models.
\newblock {\em Advances in neural information processing systems
  (NIPS/NeurIPS)}, 33:6840--6851, 2020.

\bibitem{hoogeboom2022equivariant}
Emiel Hoogeboom, Victor Garcia~Satorras, Clement Vignac, and Max Welling.
\newblock Equivariant diffusion for molecule generation in 3d.
\newblock {\em arXiv e-prints}, pages arXiv--2203, 2022.

\bibitem{huynh2020dav}
Lam Huynh, Phong Nguyen-Ha, Jiri Matas, Esa Rahtu, and Janne Heikkil{\"a}.
\newblock Guiding monocular depth estimation using depth-attention volume.
\newblock In {\em European Conference on Computer Vision (ECCV)}, pages
  581--597. Springer, 2020.

\bibitem{huynh2020guiding}
Lam Huynh, Phong Nguyen-Ha, Jiri Matas, Esa Rahtu, and Janne Heikkil{\"a}.
\newblock Guiding monocular depth estimation using depth-attention volume.
\newblock In {\em European Conference on Computer Vision}, pages 581--597.
  Springer, 2020.

\bibitem{johnston2020self}
Adrian Johnston and Gustavo Carneiro.
\newblock Self-supervised monocular trained depth estimation using
  self-attention and discrete disparity volume.
\newblock In {\em Proceedings of the IEEE International Conference on Computer
  Vision (ICCV)}, pages 4756--4765, 2020.

\bibitem{kim2022diffusion}
Boah Kim, Yujin Oh, and Jong~Chul Ye.
\newblock Diffusion adversarial representation learning for self-supervised
  vessel segmentation.
\newblock {\em arXiv preprint arXiv:2209.14566}, 2022.

\bibitem{kingma2014adam}
Diederik~P Kingma and Jimmy Ba.
\newblock Adam: A method for stochastic optimization.
\newblock {\em arXiv preprint arXiv:1412.6980}, 2014.

\bibitem{lee2019bts}
Jin~Han Lee, Myung-Kyu Han, Dong~Wook Ko, and Il~Hong Suh.
\newblock From big to small: Multi-scale local planar guidance for monocular
  depth estimation.
\newblock {\em arXiv preprint arXiv:1907.10326}, 2019.

\bibitem{lee2019big}
Jin~Han Lee, Myung-Kyu Han, Dong~Wook Ko, and Il~Hong Suh.
\newblock From big to small: Multi-scale local planar guidance for monocular
  depth estimation.
\newblock {\em arXiv preprint arXiv:1907.10326}, 2019.

\bibitem{lee2021patch}
Sihaeng Lee, Janghyeon Lee, Byungju Kim, Eojindl Yi, and Junmo Kim.
\newblock Patch-wise attention network for monocular depth estimation.
\newblock In {\em Proceedings of the AAAI Conference on Artificial
  Intelligence}, volume~35, pages 1873--1881, 2021.

\bibitem{li2022depthformer}
Zhenyu Li, Zehui Chen, Xianming Liu, and Junjun Jiang.
\newblock Depthformer: Exploiting long-range correlation and local information
  for accurate monocular depth estimation.
\newblock {\em arXiv preprint arXiv:2203.14211}, 2022.

\bibitem{li2022binsformer}
Zhenyu Li, Xuyang Wang, Xianming Liu, and Junjun Jiang.
\newblock Binsformer: Revisiting adaptive bins for monocular depth estimation.
\newblock {\em arXiv preprint arXiv:2204.00987}, 2022.

\bibitem{lin2017feature}
Tsung-Yi Lin, Piotr Doll{\'a}r, Ross Girshick, Kaiming He, Bharath Hariharan,
  and Serge Belongie.
\newblock Feature pyramid networks for object detection.
\newblock In {\em Proceedings of the IEEE Conference on Computer Vision and
  Pattern Recognition (CVPR)}, pages 2117--2125, 2017.

\bibitem{liu2023va}
Ce Liu, Suryansh Kumar, Shuhang Gu, Radu Timofte, and Luc Van~Gool.
\newblock Va-depthnet: A variational approach to single image depth prediction.
\newblock {\em arXiv preprint arXiv:2302.06556}, 2023.

\bibitem{liu2021swin}
Ze Liu, Yutong Lin, Yue Cao, Han Hu, Yixuan Wei, Zheng Zhang, Stephen Lin, and
  Baining Guo.
\newblock Swin transformer: Hierarchical vision transformer using shifted
  windows.
\newblock In {\em Proceedings of the IEEE/CVF International Conference on
  Computer Vision (ICCV)}, pages 10012--10022, 2021.

\bibitem{Silberman:ECCV12}
Pushmeet~Kohli Nathan~Silberman, Derek~Hoiem and Rob Fergus.
\newblock Indoor segmentation and support inference from rgbd images.
\newblock In {\em European Conference on Computer Vision (ECCV)}, 2012.

\bibitem{nichol2021improved}
Alexander~Quinn Nichol and Prafulla Dhariwal.
\newblock Improved denoising diffusion probabilistic models.
\newblock In {\em International Conference on Machine Learning (ICML)}, pages
  8162--8171. PMLR, 2021.

\bibitem{nwankpa2018activation}
Chigozie Nwankpa, Winifred Ijomah, Anthony Gachagan, and Stephen Marshall.
\newblock Activation functions: Comparison of trends in practice and research
  for deep learning.
\newblock {\em arXiv preprint arXiv:1811.03378}, 2018.

\bibitem{paszke2019pytorch}
Adam Paszke, Sam Gross, Francisco Massa, Adam Lerer, James Bradbury, Gregory
  Chanan, Trevor Killeen, Zeming Lin, Natalia Gimelshein, Luca Antiga, et~al.
\newblock Pytorch: An imperative style, high-performance deep learning library.
\newblock {\em Advances in neural information processing systems
  (NIPS/NeurIPS)}, 32, 2019.

\bibitem{patil2022p3depth}
Vaishakh Patil, Christos Sakaridis, Alexander Liniger, and Luc Van~Gool.
\newblock P3depth: Monocular depth estimation with a piecewise planarity prior.
\newblock In {\em Proceedings of the IEEE/CVF Conference on Computer Vision and
  Pattern Recognition (CVPR)}, pages 1610--1621, 2022.

\bibitem{qi2018geonet}
Xiaojuan Qi, Renjie Liao, Zhengzhe Liu, Raquel Urtasun, and Jiaya Jia.
\newblock Geonet: Geometric neural network for joint depth and surface normal
  estimation.
\newblock In {\em Proceedings of the IEEE Conference on Computer Vision and
  Pattern Recognition (CVPR)}, pages 283--291, 2018.

\bibitem{ranftl2021dpt}
Ren\'e Ranftl, Alexey Bochkovskiy, and Vladlen Koltun.
\newblock Vision transformers for dense prediction.
\newblock In {\em International Conference on Computer Vision (ICCV) (ICCV)},
  pages 12179--12188, 2021.

\bibitem{redmon2016you}
Joseph Redmon, Santosh Divvala, Ross Girshick, and Ali Farhadi.
\newblock You only look once: Unified, real-time object detection.
\newblock In {\em Proceedings of the IEEE Conference on Computer Vision and
  Pattern Recognition (CVPR)}, pages 779--788, 2016.

\bibitem{rombach2022high}
Robin Rombach, Andreas Blattmann, Dominik Lorenz, Patrick Esser, and Bj{\"o}rn
  Ommer.
\newblock High-resolution image synthesis with latent diffusion models.
\newblock In {\em Proceedings of the IEEE/CVF Conference on Computer Vision and
  Pattern Recognition (CVPR)}, pages 10684--10695, 2022.

\bibitem{saxena2005learning}
Ashutosh Saxena, Sung~H Chung, Andrew~Y Ng, et~al.
\newblock Learning depth from single monocular images.
\newblock In {\em Advances in neural information processing systems
  (NIPS/NeurIPS)}, volume~18, pages 1--8, 2005.

\bibitem{shao2023urcdc}
Shuwei Shao, Zhongcai Pei, Weihai Chen, Ran Li, Zhong Liu, and Zhengguo Li.
\newblock Urcdc-depth: Uncertainty rectified cross-distillation with cutflip
  for monocular depth estimation.
\newblock {\em arXiv preprint arXiv:2302.08149}, 2023.

\bibitem{silberman2012indoor}
Nathan Silberman, Derek Hoiem, Pushmeet Kohli, and Rob Fergus.
\newblock Indoor segmentation and support inference from rgbd images.
\newblock In {\em European Conference on Computer Vision}, pages 746--760.
  Springer, 2012.

\bibitem{sohl2015deep}
Jascha Sohl-Dickstein, Eric Weiss, Niru Maheswaranathan, and Surya Ganguli.
\newblock Deep unsupervised learning using nonequilibrium thermodynamics.
\newblock In {\em International Conference on Machine Learning (ICML)}, pages
  2256--2265. PMLR, 2015.

\bibitem{song2020denoising}
Jiaming Song, Chenlin Meng, and Stefano Ermon.
\newblock Denoising diffusion implicit models.
\newblock {\em arXiv preprint arXiv:2010.02502}, 2020.

\bibitem{song2021denoising}
Jiaming Song, Chenlin Meng, and Stefano Ermon.
\newblock Denoising diffusion implicit models.
\newblock In {\em International Conference on Learning Representations (ICLR)},
  2021.

\bibitem{song2019generative}
Yang Song and Stefano Ermon.
\newblock Generative modeling by estimating gradients of the data distribution.
\newblock {\em Advances in neural information processing systems
  (NIPS/NeurIPS)}, 32, 2019.

\bibitem{song2020improved}
Yang Song and Stefano Ermon.
\newblock Improved techniques for training score-based generative models.
\newblock {\em Advances in neural information processing systems
  (NIPS/NeurIPS)}, 33:12438--12448, 2020.

\bibitem{song2021scorebased}
Yang Song, Jascha Sohl-Dickstein, Diederik~P Kingma, Abhishek Kumar, Stefano
  Ermon, and Ben Poole.
\newblock Score-based generative modeling through stochastic differential
  equations.
\newblock In {\em International Conference on Learning Representations (ICLR)},
  2021.

\bibitem{tan2019efficientnet}
Mingxing Tan and Quoc Le.
\newblock Efficientnet: Rethinking model scaling for convolutional neural
  networks.
\newblock In {\em International Conference on Machine Learning (ICML)}, pages
  6105--6114. PMLR, 2019.

\bibitem{trippe2022diffusion}
Brian~L Trippe, Jason Yim, Doug Tischer, Tamara Broderick, David Baker, Regina
  Barzilay, and Tommi Jaakkola.
\newblock Diffusion probabilistic modeling of protein backbones in 3d for the
  motif-scaffolding problem.
\newblock {\em arXiv preprint arXiv:2206.04119}, 2022.

\bibitem{vaswani2017attention}
Ashish Vaswani, Noam Shazeer, Niki Parmar, Jakob Uszkoreit, Llion Jones,
  Aidan~N Gomez, {\L}ukasz Kaiser, and Illia Polosukhin.
\newblock Attention is all you need.
\newblock {\em Advances in neural information processing systems
  (NIPS/NeurIPS)}, 30, 2017.

\bibitem{wolleb2021diffusion}
Julia Wolleb, Robin Sandk{\"u}hler, Florentin Bieder, Philippe Valmaggia, and
  Philippe~C Cattin.
\newblock Diffusion models for implicit image segmentation ensembles.
\newblock {\em arXiv preprint arXiv:2112.03145}, 2021.

\bibitem{yan2021channel}
Jiaxing Yan, Hong Zhao, Penghui Bu, and YuSheng Jin.
\newblock Channel-wise attention-based network for self-supervised monocular
  depth estimation.
\newblock In {\em 2021 International Conference on 3D vision (3DV)}, pages
  464--473. IEEE, 2021.

\bibitem{yang2021transformer}
Guanglei Yang, Hao Tang, Mingli Ding, Nicu Sebe, and Elisa Ricci.
\newblock Transformer-based attention networks for continuous pixel-wise
  prediction.
\newblock In {\em Proceedings of the IEEE International Conference on Computer
  Vision (ICCV)}, pages 16269--16279, October 2021.

\bibitem{yin2019enforcing}
Wei Yin, Yifan Liu, Chunhua Shen, and Youliang Yan.
\newblock Enforcing geometric constraints of virtual normal for depth
  prediction.
\newblock In {\em Proceedings of the IEEE International Conference on Computer
  Vision (ICCV)}, pages 5684--5693, 2019.

\bibitem{yuan2021tokens}
Li Yuan, Yunpeng Chen, Tao Wang, Weihao Yu, Yujun Shi, Zi-Hang Jiang,
  Francis~EH Tay, Jiashi Feng, and Shuicheng Yan.
\newblock Tokens-to-token vit: Training vision transformers from scratch on
  imagenet.
\newblock In {\em Proceedings of the IEEE International Conference on Computer
  Vision (ICCV)}, pages 558--567, 2021.

\bibitem{Yuan_2022_CVPR}
Weihao Yuan, Xiaodong Gu, Zuozhuo Dai, Siyu Zhu, and Ping Tan.
\newblock Neural window fully-connected crfs for monocular depth estimation.
\newblock In {\em Proceedings of the IEEE Conference on Computer Vision and
  Pattern Recognition (CVPR)}, pages 3916--3925, June 2022.

\bibitem{DBLP:journals/corr/abs-2203-01502}
Weihao Yuan, Xiaodong Gu, Zuozhuo Dai, Siyu Zhu, and Ping Tan.
\newblock New crfs: Neural window fully-connected crfs for monocular depth
  estimation.
\newblock {\em CoRR}, abs/2203.01502, 2022.

\bibitem{zhang2019pattern}
Zhenyu Zhang, Zhen Cui, Chunyan Xu, Yan Yan, Nicu Sebe, and Jian Yang.
\newblock Pattern-affinitive propagation across depth, surface normal and
  semantic segmentation.
\newblock In {\em Proceedings of the IEEE/CVF Conference on Computer Vision and
  Pattern Recognition (CVPR)}, pages 4106--4115, 2019.

\end{thebibliography}
